\documentclass[journal]{IEEEtran}
\usepackage{amsmath,amsfonts}
\usepackage{algorithmic}
\usepackage{algorithm}
\usepackage{array}
\usepackage[caption=false,font=normalsize,labelfont=sf,textfont=sf]{subfig}
\usepackage{textcomp}
\usepackage{stfloats}
\usepackage{url}
\usepackage{verbatim}
\usepackage{graphicx}
\usepackage{cite}
\usepackage{enumitem}
\usepackage{booktabs}
\usepackage{multirow}
\usepackage{threeparttable}
\usepackage{epstopdf}

\makeatletter
\renewenvironment{cases}[1][l]{\matrix@check\cases\env@cases{#1}}{\endarray\right.}
\def\env@cases#1{%
  \let\@ifnextchar\new@ifnextchar
  \left\lbrace\def\arraystretch{1.2}%
  \array{@{}#1@{\quad}l@{}}}

\begin{document}

\title{Cross-View Localization via Redundant Sliced Observations and A-Contrario Validation}

\author{Yongjun Zhang$^1$,~\IEEEmembership{Member,~IEEE},
Mingtao Xiong$^1$,~\IEEEmembership{Graduate Student Member,~IEEE},\\
Yi Wan$^*$,~\IEEEmembership{Member,~IEEE},
Gui-Song Xia,~\IEEEmembership{Senior Member,~IEEE}
\thanks{This paper was produced by the National Key Research and Development Program of China, Grant No. 2024YFB3909300, the National Science Foundation of China, Grant No. 42471470. \textit{($^*$Corresponding author: Yi Wan. $^1$Both authors contributed equally to this manuscript.)}\\
\hspace*{\parindent}Yongjun Zhang, Mingtao Xiong and Yi Wan are with the School of Remote Sensing and Information Engineering, Wuhan University, Wuhan 430079, China (e-mail: zhangyj@whu.edu.cn; xiongmingtao@whu.edu.cn; yi.wan@whu.edu.cn).\\
\hspace*{\parindent} Gui-Song Xia is with the School of Artificial Intelligence, Wuhan University, Wuhan 430072, China (e-mail: guisong.xia@whu.edu.cn).
}
}


\maketitle

\begin{abstract}
Cross-view localization (CVL) matches ground-level images with aerial references to determine the geo-position of a camera, enabling smart vehicles to self-localize offline in GNSS-denied environments. However, most CVL methods output only a single observation, the camera pose, and lack the redundant observations required by surveying principles, making it challenging to assess localization reliability through the mutual validation of observational data. To tackle this, we introduce Slice-Loc, a two-stage method featuring an a-contrario reliability validation for CVL. Instead of using the query image as a single input, Slice-Loc divides it into sub-images and estimates the 3-DoF pose for each slice, creating redundant and independent observations. Then, a geometric rigidity formula is proposed to filter out the erroneous 3-DoF poses, and the inliers are merged to generate the final camera pose. Furthermore, we propose a model that quantifies the meaningfulness of localization by estimating the number of false alarms (NFA), according to the distribution of the locations of the sliced images. By eliminating gross errors, Slice-Loc boosts localization accuracy and effectively detects failures. After filtering out mislocalizations, Slice-Loc reduces the proportion of errors exceeding 10 m to under 3\%. In cross-city tests on the DReSS dataset, Slice-Loc cuts the mean localization error from 4.47 m to 1.86 m and the mean orientation error from $\mathbf{3.42^{\circ}}$ to $\mathbf{1.24^{\circ}}$, outperforming state-of-the-art methods. Code and dataset will be available at: https://github.com/bnothing/Slice-Loc.
\end{abstract}

\begin{IEEEkeywords}
Cross-view matching, camera pose estimation, robust estimation, aerial imagery, a-contrario method.
\end{IEEEkeywords}


\section{Introduction}
\IEEEPARstart{C}{ROSS-VIEW} Localization (CVL) estimates the pose of a ground-based camera by matching a query image with an aerial georeferenced image. It works by establishing visual correspondences between the two viewpoints. Aerial images are cost-effective and offer broad coverage \cite{zheng2020university}, \cite{srivastava2019understanding}, while ground images are diverse, timely, and high resolution \cite{feng2018urban}. Combining these strengths, CVL supports advanced applications like autonomous driving \cite{maddern20171}, nature disaster mapping \cite{li2025cross}, cross-view semantic segmentation \cite{ye2024sg}, and cross-view object detection \cite{sun2023cross}.

Cross-view geo-localization is generally divided into two stages \cite{ye2024coarse}: coarse and fine localization. In the coarse stage, image retrieval methods select a reference image from an extensive database (often covering an entire city). In the fine stage, the 3-Degrees-of-Freedom (3-DoF) pose of the query camera is determined using this reference image, i.e., the planar location and orientation. This paper focuses primarily on vehicle-mounted ground panoramic cameras. It is assumed that the camera pitch, roll, and shooting height relative to the ground are already known. The reference image is a geometrically processed orthophoto map with RGB channels. 

\begin{figure*}[t] 
    \centering
    \includegraphics[width=1.0\linewidth]{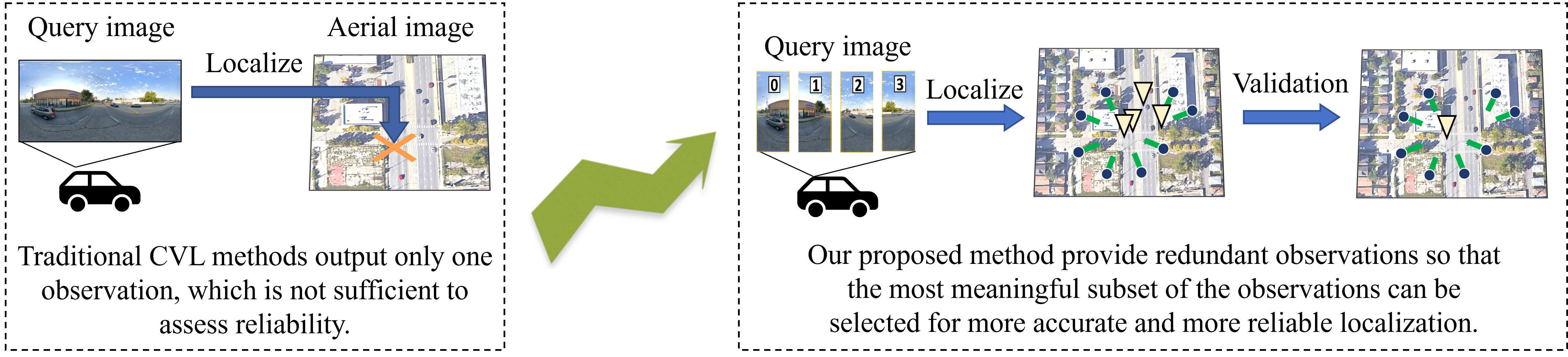}
    \caption{The left part illustrates the existing CVL pipeline, while the right part shows our improvement. By introducing redundant observations and establishing a reliability assessment mechanism, large localization errors can be detected. Determining the final localization result through mutual validation among redundant observations enhances the rigor and interpretability of CVL.}
    \label{fig:fig1}
\vspace{-0.4cm}
\end{figure*}

However, due to large differences in viewpoint, imaging modality, and capture time between the query and reference images, creating robust feature descriptors is challenging, leading to low localization accuracy. To address these challenges, several cross-view datasets have been proposed, including satellite-ground datasets (e.g., VIGOR \cite{zhu2021vigor}, CVUSA \cite{workman2015wide}, CVACT \cite{liu2019lending}, DReSS \cite{xia2024cross}), satellite-UAV datasets (e.g., University-1652 \cite{zheng2020university}), and satellite-ground vehicle datasets for fine localization (e.g., KITTI-CVL \cite{geiger2013vision}, Ford-CVL \cite{agarwal2020ford}). Research on these datasets has led to significant improvements in both coarse and fine localization. For example, Top-1 retrieval success rates have exceeded 90\% in datasets like CVUSA, CVACT, and University-1652 \cite{shen2023mccg}, \cite{ye2024cross}, and reached 75\% in VIGOR \cite{deuser2023sample4geo}. In fine localization, the localization error has been reduced to 3 m in VIGOR \cite{wang2023fine}. Despite these advancements, localization failures still occur in challenging datasets like VIGOR and DReSS, resulting in large errors \cite{xia2024cross} \cite{wang2023fine} \cite{shi2022geometry}. Moreover, as shown in Fig.~\ref{fig:fig1}, existing CVL methods lack a reliability evaluation mechanism, making it difficult to identify such failure cases.

The uncertainty in CVL has garnered attention, prompting the development of various solutions \cite{piasco2018survey}, \cite{noh2017large}. For example, C-BEV introduces a novel BEV-based retrieval module to resolve the many-to-one ambiguity in retrieval by estimating 3-DoF camera poses in the candidate references \cite{fervers2023c}. Some fine-grained cross-view localization methods predict a probability distribution over the reference image to identify predictions that potentially have large errors \cite{fervers2023uncertainty}. However, in scenes with symmetric layouts, this distribution still results in ambiguous positioning \cite{wang2023fine}, \cite{shi2022beyond}. Additionally, nearly all cross-view fine localization methods generate only a single pose, lacking redundant observations, which violates the basic surveying principles and makes it challenging to detect failed localizations via geometric consistency.

To improve pose estimation accuracy and reliability, we propose Slice-Loc, a reliability-aware method that predicts the locations of sub-scenes depicted in a query image for 3-DoF ground camera pose estimation while evaluating result usability. Unlike commonly used end-to-end approaches, Slice-Loc operates in two stages. First, the ground-level image is divided into sub-images, and then the 3-DoF pose of each sub-image is estimated to provide redundant observations. Then, error detection is performed by traversing sampled inlier sets in a RANSAC-based workflow, and the inlier poses are merged to compute the ground-level camera pose. We propose a novel model to approximate the NFA for CVL. This formula probabilistically leverages the results of robust estimation, jointly considering the geometric errors of the 3-DoF poses from sliced images and the number of inliers. If the NFA exceeds a set threshold, the localization is considered unreliable. 

To summarize, our main contributions are as follows:
\begin{itemize}
\item A novel two-stage camera pose estimation framework for CVL, called Slice-Loc, is proposed. Our method slices the ground image into multiple sub-images and independently localizes each slice to provide redundant observations for self-validation in CVL. Slice-Loc integrates robust estimation with reliability assessment, achieving an automated and rigorous CVL process. Experiments on the VIGOR and DReSS datasets demonstrate that our approach generalizes to new urban areas and far outperforms state-of-the-art methods. 
\item A probabilistic criterion is derived to evaluate the geometric consistency of a set of 3-DoF poses from sliced images. Leveraging this criterion, we develop a threshold-free method, OSA-CVL, which estimates the camera pose by merging the inlier slice poses. Based on a-contrario theory, a model is proposed to evaluate the meaningfulness of a CVL result, which can efficiently identify the failed localizations to prevent large errors.
\item A fine-grained geo-tagged CVL dataset, DReSS-D, is introduced to provide pixel-level ground truth labels for training models. Each ground panoramic image is paired with a depth map, enabling every pixel in the query image to be projected into the reference image space. To our knowledge, DReSS-D is the first academic cross-view dataset to offer pixel-level correspondences, enabling denser supervision beyond traditional camera-pose labels.
\end{itemize}

The remainder of this paper is structured as follows: Section \ref{related work} provides a comprehensive review of related work, including retrieval-based cross-view geo-localization, fine-grained cross-view localization, and gross error detection methods. Section \ref{dataset} introduces the proposed DReSS-D dataset and the pixel-level projection of the ground image onto the reference image via a depth map. Section \ref{method} outlines the proposed method, Slice-Loc. Experimental results and discussions are presented in Sections \ref{experiment} and \ref{discussion}, respectively. Finally, Section \ref{conclusion} concludes the paper with an analysis of the method’s advantages and limitations.

\section{Related Work}
\label{related work}
In this section, we review the work related to cross-view camera pose estimation and gross error detection.
\vspace{-10pt}
\subsection{Retrieval-Based Cross-View Geo-Localization}
Cross-view image retrieval methods have shown significant progress in recent years. This technology efficiently provides both a coarse geographic location and a reference image for a query ground-level image by comparing the similarity of image features. Extracting feature descriptors that are robust to large view differences for the same scene is the key to cross-view image retrieval. Earlier studies used manually designed operators to extract features from both query and reference images \cite{hays2008im2gps}, \cite{lin2013cross}, \cite{fu2017fast}. With the development of deep learning, researchers have since developed Siamese neural networks specifically tailored for cross-view visual retrieval tasks. These networks use two branches to independently learn visual features from ground-level and aerial images \cite{lin2015learning}, \cite{hu2018cvm}.

Generating feature representations that bridge the domain gap between aerial and ground views is a key challenge in cross-view retrieval. CVM-Net \cite{hu2018cvm} addresses this by integrating a NetVLAD layer with a Siamese neural network to aggregate local features into a view-invariant global descriptor. In \cite{liu2019lending}, the discriminative ability of deep features is enhanced by explicitly encoding the orientation of each pixel in the image. CVFT \cite{shi2020optimal} utilizes a feature transfer technique to map deep features into a new domain space, ensuring consistent feature distributions across domains. In \cite{wang2021each}, a local pattern network is proposed to capture scene context information, forcing the model to focus on important but often overlooked features. Additionally, guiding deep learning models to automatically prioritize image regions relevant to retrieval has proven effective in improving accuracy. Since the advent of the Transformer \cite{vaswani2017attention}, its powerful feature aggregation capabilities have been leveraged for cross-view retrieval tasks. L2LTR \cite{yang2021cross} uses self-attention and position encoding to combine local visual features with geometric information into a unified descriptor, while TransGeo \cite{zhu2022transgeo} introduces an attention-guided non-uniform cropping method that removes irrelevant image sections, boosting retrieval efficiency and recall compared to CNN-based methods. Sample4Geo \cite{deuser2023sample4geo} further enhances feature representation through a contrastive learning framework with a hard negative sample strategy. Inspired by this work, recent methods have adopted finer prior supervisions \cite{xia2024cross} and multi-query fusion strategies \cite{wu2024crossviewimagesetgeolocalization} to enhance the discriminative capability of image feature representations.

Geometric transformation can warp an image from one view to another, reducing the view difference between the two types of images and effectively increasing the retrieval recall rate. For example, SAFA \cite{shi2019spatial} applies a scene-independent polar coordinate transformation to convert aerial images into ground-level images. In contrast, \cite{li2023multi} uses a reverse polar transformation to generate a top-down view of ground images. However, without depth information, these transformations can introduce additional distortions, negatively impacting retrieval performance. 

\vspace{-5pt}
\subsection{Fine-Grained Cross-View Localization}
Fine-grained cross-view localization seeks to determine a ground-level sensor's position and orientation relative to a reference image obtained via retrieval or approximate GNSS. Using matching or regression techniques, fine localization can yield more accurate results than coarse localization methods. \cite{zhu2021vigor} introduced a city-level dataset for fine localization, emphasizing the challenges and importance of fine localization beyond one-to-one retrieval.

Initially, global image features were primarily used to predict camera pose. For instance, in \cite{hou2022road}, the road is treated as the main recognition object, and the camera’s position is regressed by aligning the road masks from the ground and aerial images. In \cite{hu2022beyond}, the problem of estimating the camera’s rotation angle and a corresponding evaluation metric are defined. Recent studies have shown that local features offer greater advantages in fine localization. In \cite{xia2022visual}, dense local features are extracted from the reference image, and a distribution heatmap is generated through similarity comparison and feature decoding. SliceMatch \cite{lentsch2023slicematch} produces a set of geometry-aware aerial descriptors to determine the camera pose from candidate locations. CCVPE \cite{xia2023convolutional} extracts local features at multiple scales and employs a novel rolling descriptor method to reduce rotational differences between reference and query images, surpassing previous state-of-the-art methods in both accuracy and efficiency.

Geometry-based transformations are also applied in fine localization to reduce the domain gap between aerial and ground views. In \cite{shi2022beyond}, satellite features are projected into the ground-view feature space by simulating the ground camera's imaging process, and the LM optimization algorithm is used to polish the camera pose iteratively. GGCVT \cite{shi2023boosting} introduces a geometrically guided transformer to map ground features to the satellite perspective. Additionally, \cite{wang2024view} proposes the T2GA module, which leverages reliable off-ground information to improve positioning accuracy under challenging scenes such as visual occlusion. \cite{fervers2023uncertainty} generates bird’s eye view (BEV) features of ground image through Vit \cite{dosovitskiy2020image}, and compares them with the reference features to derive the probability distribution of the query camera.

Given the poor generalization of cross-view fine localization methods in new target regions and the challenges in obtaining accurate ground truth, researchers have turned to weakly supervised approaches. For example, in \cite{xia2024adapting}, a weakly supervised learning method based on knowledge self-distillation guides a student model with pseudo-labels generated by a teacher model. In another study \cite{shi2024weakly}, noisy GPS data are used to identify positive and negative samples to form pseudo query-reference pairs, and contrastive learning is applied to train the network.

In summary, while fine-grained cross-view localization has significantly improved accuracy and generalization, incorrect positioning still occurs. Moreover, most methods rely on end-to-end deep learning to predict camera poses, making it difficult to assess the reliability of the results directly from the inference process. This limitation can significantly hinder the practical application of cross-view localization.

\vspace{-10pt}
\subsection{Gross Error Detection}
In the visual localization, correspondences between query and reference images are established by extracting and matching visual features to compute the relative pose \cite{miao2024survey}, \cite{liu2025render}. When redundant observations are available, robust estimation methods leverage geometric constraints to filter out incorrect correspondences and refine the output pose \cite{li2021robust}.

RANSAC-type methods \cite{fischler1981random}, \cite{mishkin2015mods} distinguish inliers from outliers by randomly sampling a geometric model to measure the error. These methods can handle cases with over $50$\% outliers \cite{li2023qgore}. In different application scenarios, selecting an appropriate geometric model is crucial to detect gross errors. For instance, when using pinhole cameras, corresponding points satisfy the epipolar geometry constraint \cite{moisan2004probabilistic}. In contrast, if the camera's optical center is fixed or the viewed scene is planar, a homography transformation can be used for registration \cite{moisan2012automatic}. In some applications, flexible geometric constraints can better detect gross errors. For example, \cite{wan2017p2l} proposes an epipolar line segment model that uses prior elevation data to reduce mismatches between satellite images. In ground-aerial image matching, the reference orthophoto is treated as a vertical parallel projection, with the query camera aligned parallel to the ground, so that the epipolar line in the query image is perpendicular to the ground \cite{wang2023fine}, \cite{wang2023satellite}.

The a-contrario RANSAC method called ORSA is proposed to detect rigid point matches \cite{moisan2004probabilistic}. In the context of a-contrario theory, all events are assumed to occur under the null hypothesis, and the expected number of occurrences of the observation results (referred to as NFA) is calculated to assess whether the result is reliable. In the ORSA method, the optimal subset is obtained through multiple random samplings. The NFA is calculated based on the number of elements in the subset and the maximum geometric error, and it is then used to determine whether the matched result is available. This NFA-based reliability determination method was initially proposed by \cite{desolneux2000meaningful}. Under the null hypothesis, perfect coincidence results are highly unlikely, which is the desired outcome. This theory is commonly applied to detect geometric structures in images, such as mosaic parts \cite{bammey2023contrario}, line segments \cite{von2008lsd}, junctions \cite{xia2014accurate}, change areas \cite{robin2010contrario}, and tiny objects \cite{deep_nfa}.

\section{Problem and Dataset Construction}
\label{dataset}
First, we outline our problem statement about CVL dataset. Subsequently, we introduce specific details of the proposed dataset. Finally, we present a method for establishing pixel-level correspondences based on our dataset.

\vspace{-10pt}
\subsection{Problem Statement}
Currently, researchers have proposed several CVL datasets, such as VIGOR \cite{zhu2021vigor}, DReSS \cite{xia2024cross}, CVUSA \cite{workman2015wide}, CVACT \cite{liu2019lending}, and KITTI-CVL \cite{geiger2013vision}. In these datasets, the query image is a ground-level street view image, while the reference image is a satellite image covering the same scene, forming an image pair. Each image pair provides only coarse scene-level supervision (i.e., the camera’s 3DoF pose). However, we employ an indirect localization method that requires the sub-scene’s location (rather than the camera's location) as the ground truth for training the localization model. Inspired by image matching datasets like MegaDepth \cite{li2018megadepth} and ScanNet \cite{dai2017scannet}, we use absolute depth values to determine the scene’s location in the reference image. It is important to note that depth data is used solely to generate ground truth and is not involved during the localization process.

\vspace{-10pt}
\subsection{Data Collection}
Building on DReSS \cite{xia2024cross}, we introduce the DReSS-D dataset, which includes ground and aerial images from six cities across six continents: Sydney, Chicago, Johannesburg, Tokyo, Rio, and London, offering diverse scene variations. The images of each city are evenly distributed, covering an area of over $400$ $km^2$. The depth maps are obtained via the Google Street View Static API, which provides $360$-degree depth images rendered from $3$D models in Google Earth \cite{anguelov2010google}. The depth maps mainly contain the depth information of buildings and roads relative to the camera's position. Following the setting in \cite{zhu2021vigor}, \cite{shi2022beyond}, the reference image is an orthophoto map with a resolution of approximately $0.11$ meters and a size of $1280$. The center of the reference image corresponds to the GPS coordinates of the ground camera.

\vspace{-10pt}
\subsection{Pixel-Level Correspondence}
\begin{figure}[t!] 
    \centering
    \includegraphics[width=1.0\linewidth]{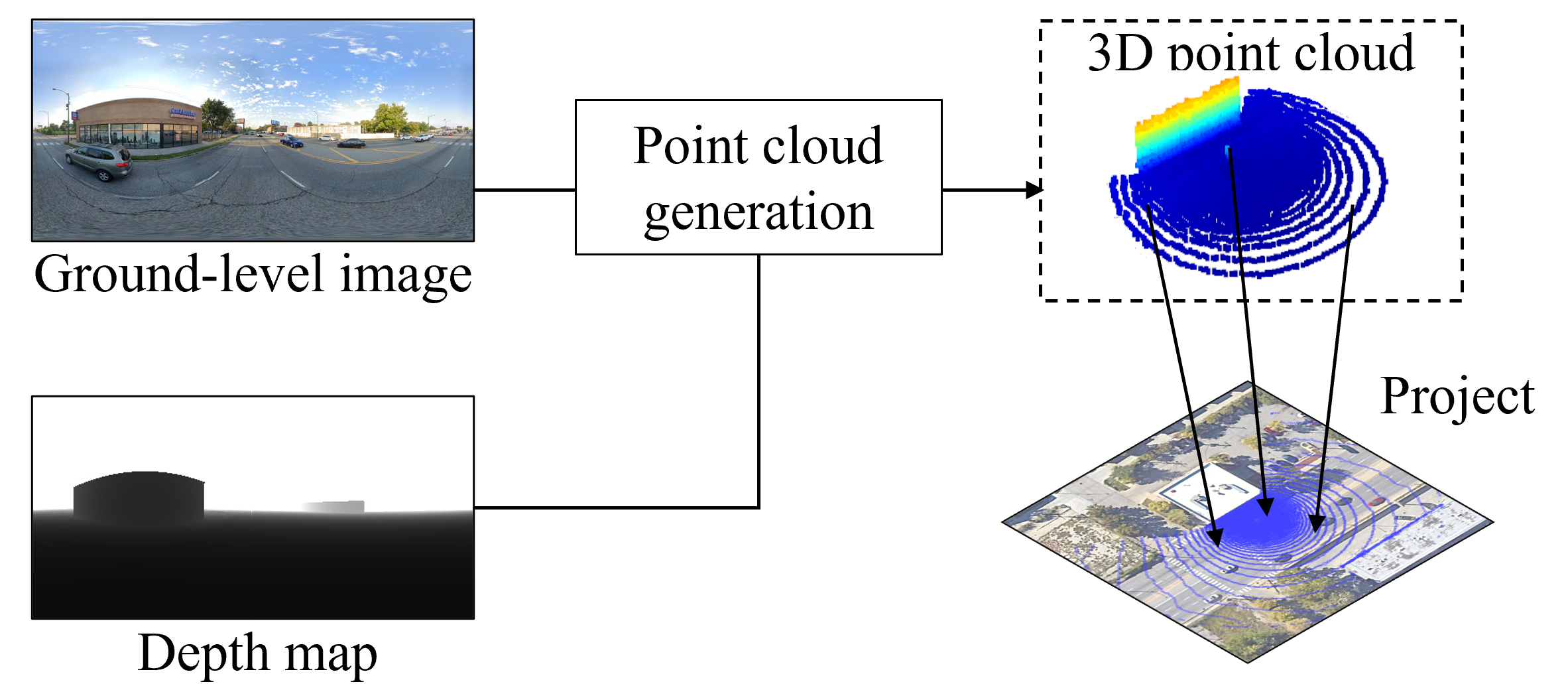}   
    \caption{Pixel-level correspondence. Using the depth map, a $3$D point cloud of the panoramic image in object space is generated and then projected onto the reference image using the depth map.}
    \label{fig:fig2}
\vspace{-0.4cm}
\end{figure}

As shown in Fig.~\ref{fig:fig2}, the DReSS-D dataset establishes a pixel-level one-to-one correspondence between the ground panoramic image and the satellite image. For pixel $i$ in the ground image, the world coordinates of $i$ can be determined using the panoramic camera imaging model and the depth map. The pixel coordinates $(x_i,y_i)$ in the ground query image are first converted to angular coordinates $(\varphi_i,\omega_i)$:
\begin{align}
    \varphi_i=2 \pi \frac{x_i}{W}, \omega_i=\pi \frac{y_i}{H},
    \label{eq:1}
\end{align}
where \(W\) and \(H\) are the width and height of the panoramic image. These coordinates are then projected into the world coordinate system based on the depth:
\begin{align}
\begin{bmatrix}X_i\\Y_i\\Z_i\end{bmatrix}
=\begin{bmatrix}X_c \\Y_c \\Z_c\end{bmatrix}+
\operatorname{depth}\left(x_i, y_i\right)
\begin{bmatrix}
\sin \left(\omega_i\right) * \sin \left(\varphi_i\right) \\
\sin \left(\omega_i\right) * \cos \left(\varphi_i\right) \\
\cos \left(\omega_i\right)
\end{bmatrix},
\end{align}
where $\begin{bmatrix}X_i,Y_i,Z_i\end{bmatrix}^{T}$ is the $3$D world coordinates of pixel $i$, $\begin{bmatrix}X_c,Y_c,Z_c\end{bmatrix}^{T}$ is the world coordinates of ground camera, $\operatorname{depth}\left(x_i, y_i\right)$ represents the depth value of $i$. The $3$D coordinates of pixel $i$ are projected to the reference image space using the geo-coordinate transformation parameters:
\begin{align}
x_i^{(s)}&=\left(X_i-\operatorname{lon}_{t l}\right) \frac{W_s}{lon_{tl}-lon_{br}}, \\
y_i^{(s)}&=\left(Y_i-lat_{tl}\right) \frac{H_s}{lat_{br}-lat_{tl}},
\end{align}
where $({lon_{tl}},{lat_{tl}})$ represents the latitude and longitude of the pixel in the upper-left corner of the reference image, $({lon_{br}},{lat_{br}})$ is the latitude and longitude of the pixel in the lower-right corner. ${W_s}$ and ${H_s}$ denote the reference image's width and height, respectively.

\section{Proposed Method}
\label{method}
\begin{figure*}[t!] 
    \centering
    \includegraphics[width=1.0\linewidth]{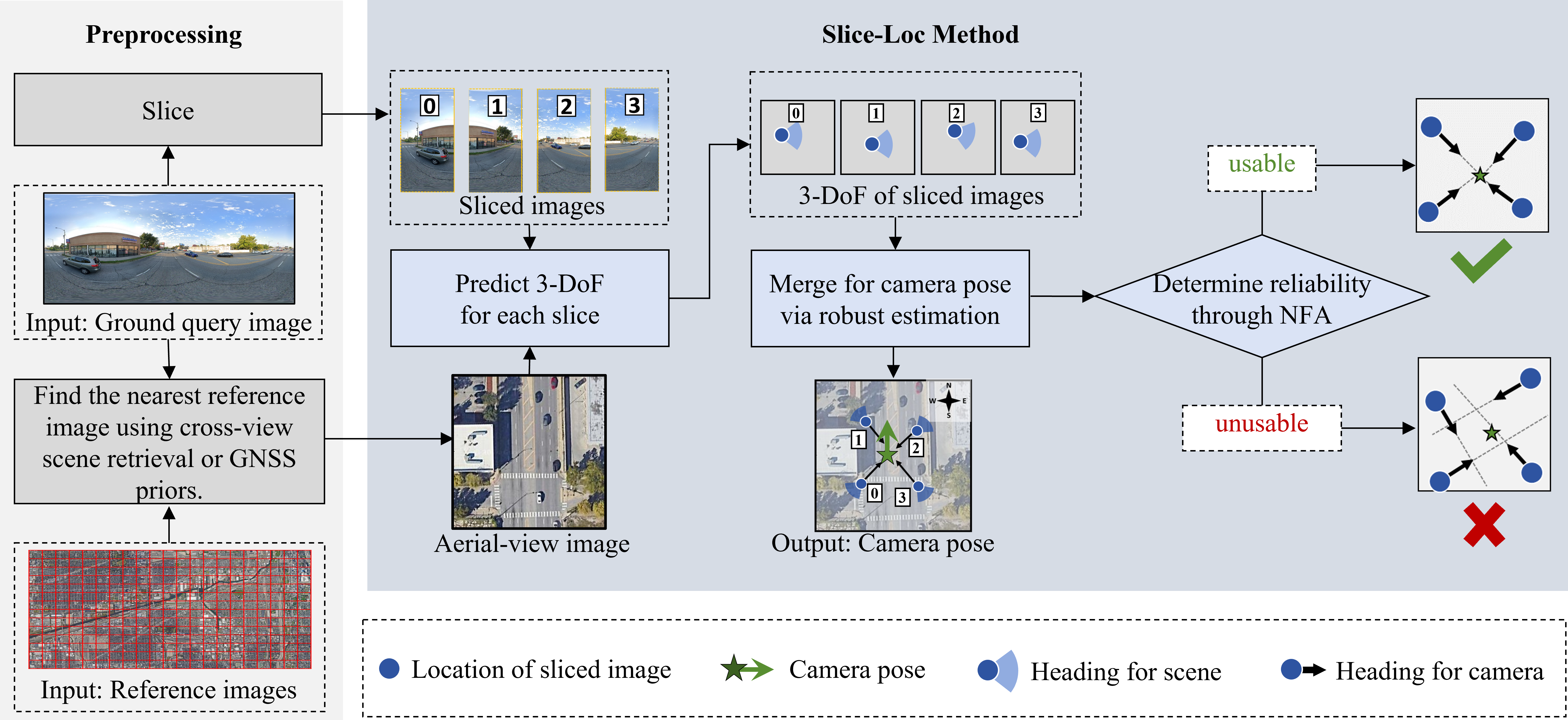}
    \caption{Flowchart of the proposed Slice-Loc method. In the preprocessing stage, the Horizontal Field-of-View of the ground query image is uniformly divided into ‘slices’, and the aerial view image is selected from the reference image database through retrieval or GNSS information. In the Slice-Loc method, the 3-DoF of each sliced image is determined in the reference image as a redundant observation. Using these 3-DoFs and their geometric relationships, the ground camera pose is estimated in a robust estimation pipeline. Finally, the NFA of the localization is estimated to predict the validity.}
    \label{fig:fig3}
\vspace{-0.4cm}
\end{figure*}

In this section, we introduce our proposed method: Slice-Loc. As illustrated in Fig.~\ref{fig:fig3}, the Slice-Loc method consists of three steps: 1) determining the 3-DoF pose of the sliced images; 2) estimating the camera pose; and 3) determining the reliability of the estimation. 

Specifically, in step 1, the horizontal field-of-view (HFoV) of the ground query image is uniformly divided into "slices" and the 3-DoF of each sliced image is determined using the reference image. In step 2, based on a defined geometric metric, we introduce a measure of set rigidity to differentiate inliers from outliers among these 3-DoFs while simultaneously estimating the camera pose. Finally, the NFA of the pose estimation process is computed using a-contrario theory to determine whether the camera pose is reliable.

\textit{\textbf{Problem Formulation:}} Given an image pair $\{I_G, I_R\}$, where $I_G$ is a ground query image and $I_R$ is a reference aerial-view image covering the same scene, our goal is to estimate the relative translation and rotation between $I_G$ and $I_R$, as well as a reliability indicator:
\begin{align}
\left(\hat{\boldsymbol{p}}_c, \hat{s}\right)=f\left(I_G, I_R\right),
\end{align}
Where $\hat{\boldsymbol{p}}_c=(\boldsymbol{p}_c,\phi_c)$ is the 3-DoF pose of query camera, and $\boldsymbol{p}_c=(x_c,y_c)$ is the location, $\phi_c$ represents the angle between the camera's orientation and true north: $0^{\circ}$ indicates the camera is facing true north, and the angle increases in the clockwise direction. The scalar $\hat{s}$ determines the validity of the camera pose $\hat{\boldsymbol{p}}_c$ by comparing it with a well-defined threshold $\tau$.

\vspace{-10pt}
\subsection{Sliced Image Pose Estimation}
\begin{figure*}[t!] 
    \centering
    \includegraphics[width=1.0\linewidth]{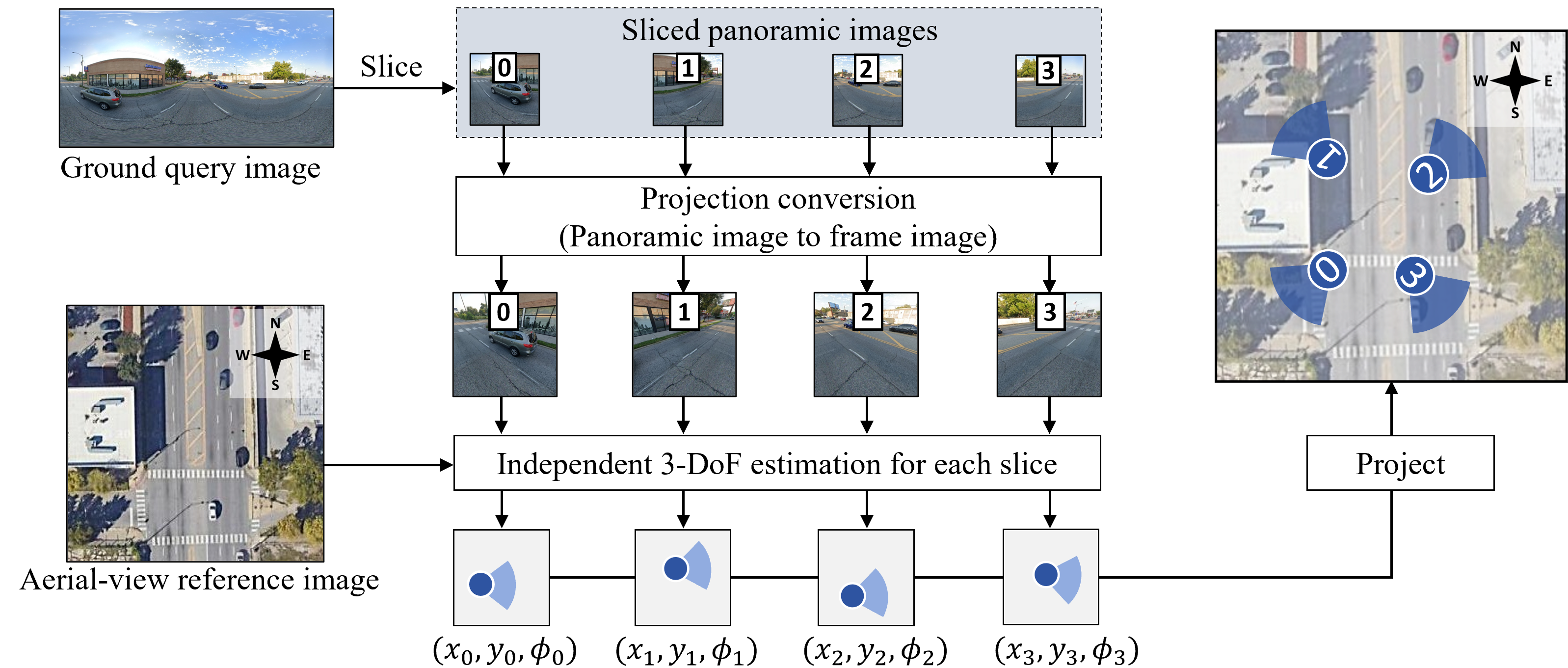}
    \caption{Flowchart of 3DoF estimation for sliced images. The relative 3-DoF poses between each sliced image and the reference image are then determined. Finally, these 3-DoF poses are projected back into the original image space.}
    \label{fig:fig4}
\vspace{-0.4cm}
\end{figure*}

Given a ground image $I_G$, unlike previous methods \cite{shi2022beyond}, \cite{xia2023convolutional} that directly predict the camera pose, we first predict the 3-DoF poses of sliced images denoted as set $S_G=\{\hat{\boldsymbol{p}}_i\}^n$, and then use $S_G$ to estimate the camera's pose $\hat{\boldsymbol{p}}_c$. Each sliced pose $\hat{\boldsymbol{p}}_i=(\boldsymbol{p}_i,\phi_i)$ consists of two components: a scene coordinates $\boldsymbol{p}_i$ and a scene orientation angle $\phi_i$. The scene coordinate $\boldsymbol{p}_i$ represents the position of the scene depicted in the query image, relative to the reference image, while the scene orientation angle $\phi_i$ denotes the clockwise angle between the direction from the scene position to the camera position and the true north direction. The prediction process for $S_G$ is shown in Fig.~\ref{fig:fig4}.

\begin{figure}[t!] 
    \centering
    \includegraphics[width=1.0\linewidth]{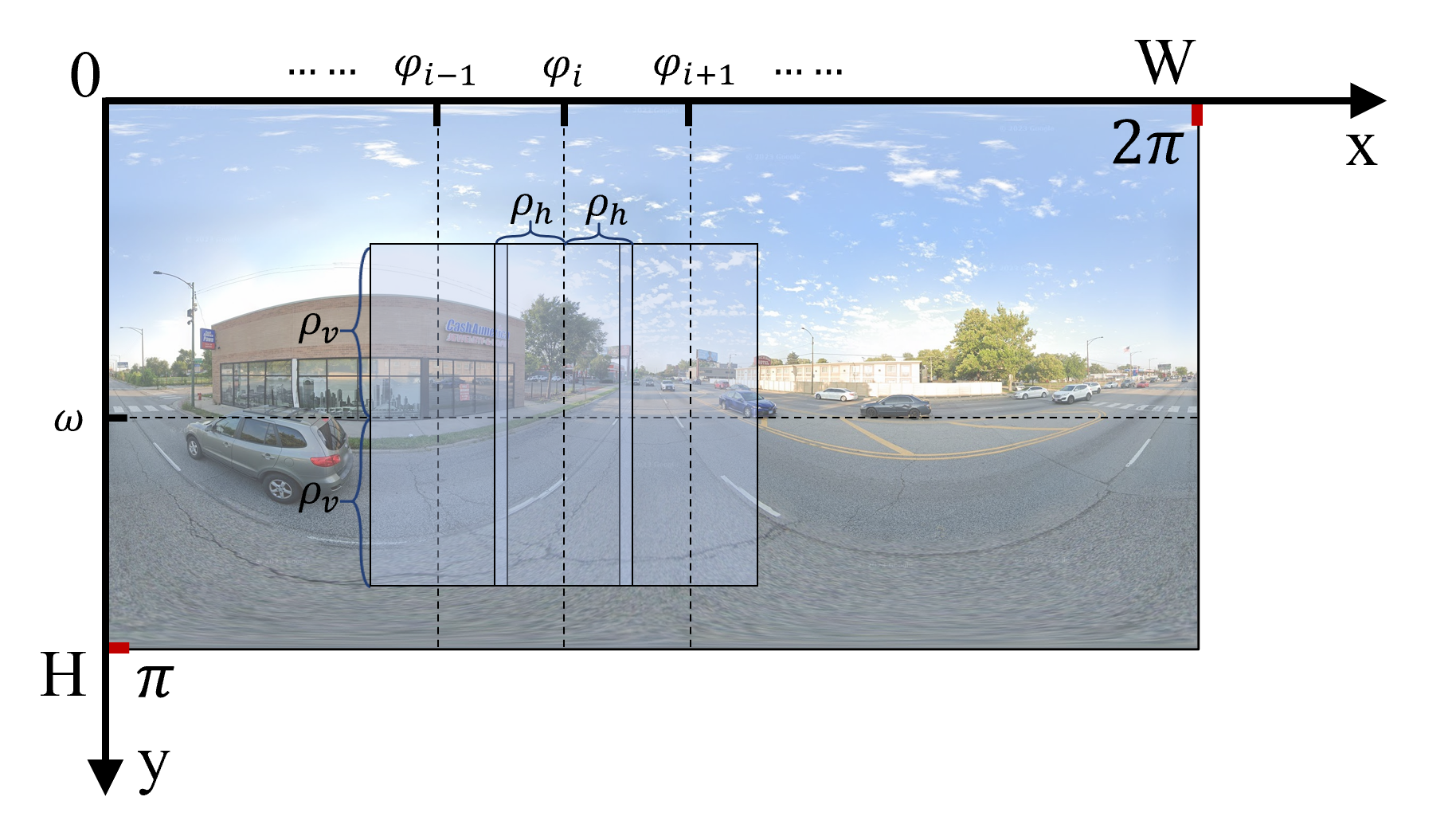}   
    \vspace{-15pt}
    \caption{Illustration of panoramic image view range and sliced image view range. The centers of the sliced images are uniformly distributed along the x-axis and share the same y-coordinate.}
    \label{fig:fig5}
\vspace{-0.4cm}
\end{figure}

\textit{Slicing Query Image:} The sliced images have the same size, with the same field of view. For a sliced image $G_i$, its horizontal field of view (HFoV) is $2\rho_h$, and the vertical field of view (VFoV) is $2\rho_v$. For each $G_i$, the VFoV center $\omega_i=0.75\pi$ against the original panoramic image. The HFoV center $\varphi_i$ of sliced images are evenly distributed in $[0,2\pi]$, i.e. $\varphi_i=2\pi\cdot i/n,i=1\cdots n$, $n$ is number of sliced images. The HFoV and VFoV of the sliced image are shown in Fig.~\ref{fig:fig5}. To reduce differences in imaging modes, the sliced panoramic-type image is resampled into the frame-type image by transforming the image projection system.

\textit{Rotating Reference Image:} Before predicting the 3-DoFs of the sliced images, to reduce the search range of the rotation angle, the reference image is rotated according to the center angle $\varphi_i$ of the sliced image’s HFoV and the orientation prior of the ground camera. The rotation ensures that the output rotation angles for all sliced images are within the same range (set to $[45^{\circ},135^{\circ}]$ in this paper). After obtaining the 3-DoFs set $S_G$, rotation restoration is applied to project the poses in $S_G$ into the original image space of the reference image $I_R$.

\textit{Supervision for Training:} We use the deep learning model from CCVPE \cite{xia2023convolutional} to predict the 3-DoF of each sliced image. Fine-grained cross-view localization methods generally fall into two categories: feature-based and geometric-based approaches. While geometric-based methods use transformations during inference and training, our approach relies on geometric patterns to identify gross errors and determine reliability, using feature-based methods better ensures the independence between the poses in $S_G$. Unlike the original CCVPE method, which uses camera position as supervision, we use scene position. This is more realistic because the camera position in the reference image often deviates from the true scene position, which can mislead the model into matching features from incorrect locations during training. We verify this effect through experiments (see Section \ref{sec:abl_stu}).

The scene coordinates of the sliced image are obtained by projecting each pixel of the sliced image into the reference image and calculating the mean of all the projected coordinates. Pixels in regions with invalid depth, such as the sky part, are excluded from the calculation. For the sliced image $G_i$ its scene position $\boldsymbol{p}_i=(x_{sce}^{(i)},y_{sce}^{(i)})$ is obtained by the following formula:
\begin{align}
\setlength\abovedisplayskip{10pt}
\setlength\belowdisplayskip{10pt}
x_{s c e}^{(i)}&=\operatorname{mean}\left(\left\{x_k^{(i)} \mid k \in G_i \text { and }\operatorname{depth}(k)<255\right\}\right), \\
y_{s c e}^{(i)}&=\operatorname{mean}\left(\left\{y_k^{(i)} \mid k \in G_i \text { and }\operatorname{depth}(k)<255\right\}\right),
\end{align}
where $(x_k^{(i)},y_k^{(i)})$ represents the projection coordinates of pixel $k$ in the reference image, obtained through projection using the depth map.

\vspace{-10pt}
\subsection{Measuring Rigidity}
To filter out the inaccurate poses in $S_G$, we define a metric that describes the geometric rigidity of the sliced images' poses, and propose a probabilistic method to compute the metric.

\textit{Panoramic Geometric Constraint:} The panoramic image can be modeled as a sphere, with light from all directions converging onto a camera at its center. This imaging process is projected onto the reference map, analogous to multiple lines in a 2D plane intersecting at a single point. This process establishes the geometric constraint for the poses in $S_G$.

\begin{figure}[t!] 
    \centering
    \includegraphics[width=1.0\linewidth]{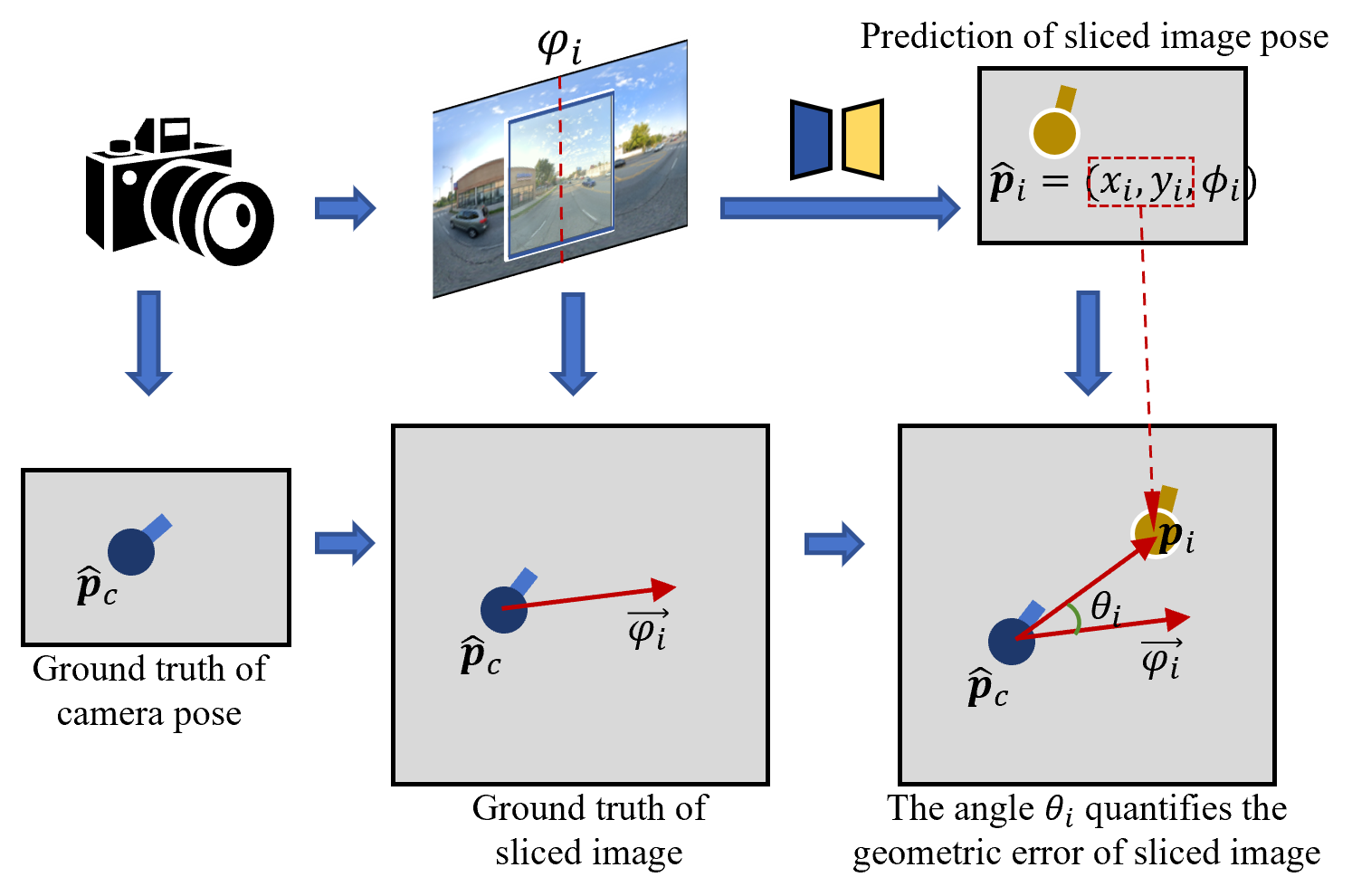}   
    \caption{Illustration of geometric error for the sliced image. Given the camera’s pose, the geometric error is determined by the sliced image’s pose and the HFoV center.}
    \label{fig:fig6}
\vspace{-0.4cm}
\end{figure}

As shown in Fig.~\ref{fig:fig6}, for the pose $\hat{\boldsymbol{p}}_i$ of sliced image $G_i$, the geometric error is defined as the angle between the vector formed by the sliced image position $p_i$ and the camera position $p_c$, and the central imaging ray of $G_i$:
\begin{align}
    \theta_i(\hat{\boldsymbol{p}}_i)=<\overrightarrow{\boldsymbol{p}_c \boldsymbol{p}_i},\overrightarrow{\varphi_i}>,
    \label{eq:geo_err}
\end{align}
where ${<}\cdot{>}$ represents the angle between two vectors, $\overrightarrow{\boldsymbol{p}_c \boldsymbol{p}_i}$ is the vector from camera position $\boldsymbol{p}_c$ to sliced image position $\boldsymbol{p}_i$. Vector $\overrightarrow{\varphi_i}$ is the central imaging ray from $\boldsymbol{p}_c$ to the center of $G_i$, and is determined by the camera pose $\hat{\boldsymbol{p}}_c$ and the HFoV center $\varphi_i$ of sliced image $G_i$. Both $\overrightarrow{\boldsymbol{p}_c \boldsymbol{p}_i}$ and $\overrightarrow{\varphi_i}$ are in the image space of reference image $I_R$. We assume the centroid of $G_i$ aligns with its center in the reference map, meaning that ideally $\theta_i(\hat{\boldsymbol{p}}_i)$ is 0.

The camera position $\boldsymbol{p}_c$ is estimated by the subset of $S_G$. For a subset $S_G' \subseteq S_G$, $\boldsymbol{p}_c$ is the point that achieves minimum geometric error for $S_G'$:
\begin{align}
\boldsymbol{p}_c=\underset{\boldsymbol{p}_c}{\operatorname{argmin}}\left(\sum_{i=1}^k \theta_i\left(\hat{\boldsymbol{p}}_i\right)\right), \boldsymbol{p}_c \in I_R, \hat{\boldsymbol{p}}_i \in \mathcal{S}_G^{\prime},
\label{eq:came_loc}
\end{align}
where $k$ is the number of elements in $\mathcal{S}_G^\prime$, and satisfies $k\geq2$. However, the camera orientation angle $\phi_c$ is the average of the orientation angles of each sliced pose $\left\{\phi_i\middle| i=1,\ldots,n\right\}$ as follows:
\begin{align}
\phi_c=\operatorname{mean}(\{\phi(\hat{\boldsymbol{p}}_i)\mid\hat{\boldsymbol{p}}_i \in \mathcal{S}_G\}),
\end{align}
here, $\phi(\cdot)$ represents the orientation angle of pose ${\hat{\boldsymbol{p}}}_i$.

\textit{Geometric Rigidity Evaluation:} In this part, we use a probability-based formula to measure the geometric quality of the 3-DoF pose set $\mathcal{S}_G$. Consistent with previous a-contrario work \cite{moisan2004probabilistic}, our measure process is carried out under the null hypothesis $\mathcal{H}_0$. In the null hypothesis $\mathcal{H}_0$, the query image is assumed to consist entirely of meaningless white noise, making it challenging to extract visual cues for localization. As a result, the localization outcomes follow a specific distribution.

\begin{figure}[!t]
\centering
\subfloat[]{\includegraphics[width=0.48\linewidth]{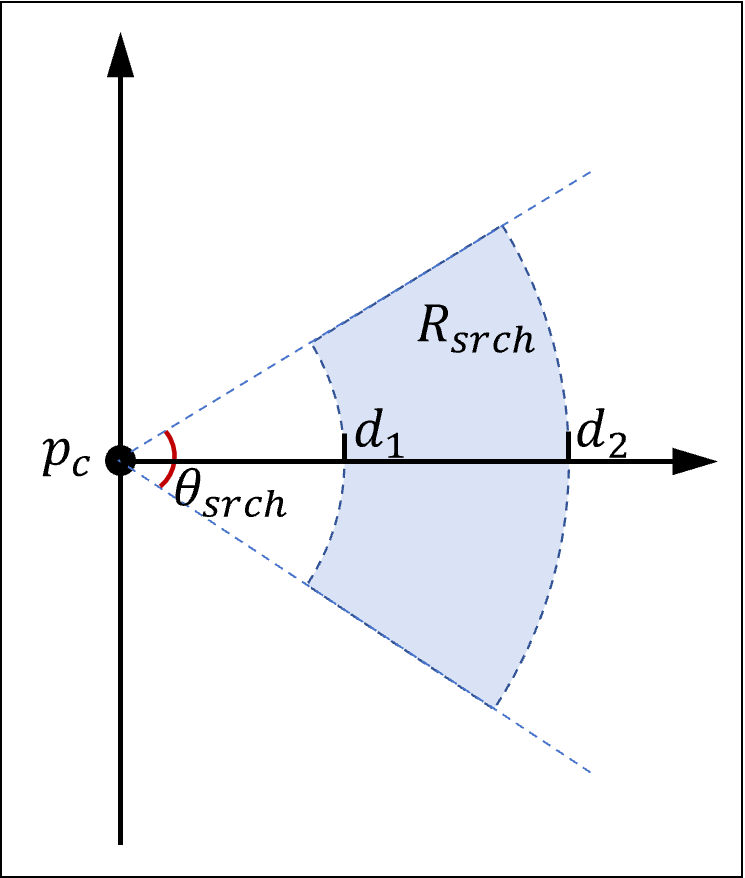}%
\label{fig:fig7_a}}
\hfil
\subfloat[]{\includegraphics[width=0.48\linewidth]{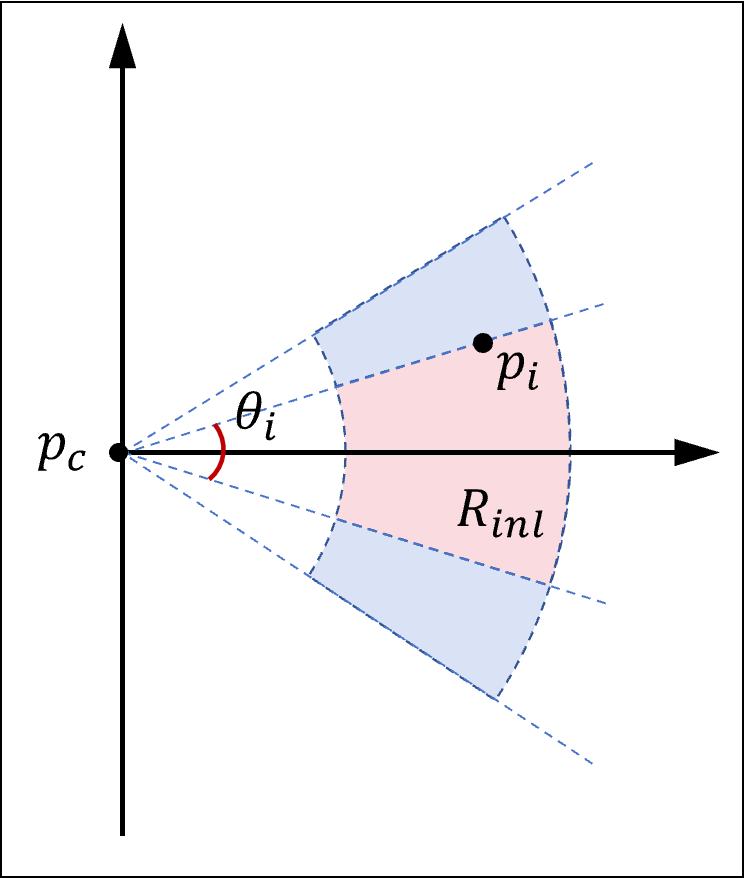}%
\label{fig:fig7_b}}
\caption{The definition of search region and inlier region. (a) The search region $R_{srch}$ is determined by the a priori error, where the sliced 3-DoF observation may occur. (b) The inlier region $R_{inl}$ is determined by the sliced 3-DoF observation; any other observations within this region will incur smaller errors. $d_1$ and $d_2$ are the inner and outer radii of the annular sector, shared by the search region and the inlier region. $\theta_{srch}$ and $\theta_i$ are the central angle of the search region and inlier region, respectively. $p_i$ is the location point of the sliced image.}
\label{fig:fig7}
\vspace{-0.4cm}
\end{figure}

Under the null hypothesis $\mathcal{H}_0$, when localizing a sliced image $G_i$, only the camera pose, the HFoV of $G_i$, and the camera orientation prior are known. Therefore, the search region $R_{srch}$ is defined as an annular sector region on the reference image, with the camera position $\boldsymbol{p}_c$ as the origin, as shown in Fig.~\ref{fig:fig7}(a). The inner and outer radii of the annular sector, $d_1$ and $d_2$, are determined by the ground camera's height, pitch angle, and VFoV. The angle $\theta_{srch}$ is determined by the HFoV of $G_i$ and the camera's orientation prior. For clarity, the positive x-axis is aligned with the central vector $\overrightarrow{\varphi_i}$.

As shown in Fig.~\ref{fig:fig7}(b), for a sliced image $G_i$, the inlier region $R_{inl}$ is defined as an annular sector region with angle $\theta_i$. According to the geometric error defined in (\ref{eq:geo_err}), the sliced image pose ${\hat{\boldsymbol{p}}}_j$ has a better geometric consistency than ${\hat{\boldsymbol{p}}}_i$, when the location point $\boldsymbol{p}_j$ of ${\hat{\boldsymbol{p}}}_j$ is in $R_{inl}$

To simplify the expression, we define the event $e_i(\mathcal{\vartheta})$ as the observation that angle $\theta_i$ is less than $\vartheta$. Under the null hypothesis $\mathcal{H}_0$, the point $p_i$ is distributed within the search region $R_{srch}$, and the probability of event $e_i(\mathcal{\vartheta})$ is as follows:
\begin{equation}
\begin{aligned}
Q(\vartheta) &:=\operatorname{Prob}\left(e_i(\vartheta) \mid \mathcal{H}_0\right) \\
& =\operatorname{Prob}\left(\theta_i \leq \vartheta \mid \mathcal{H}_0\right) \\
& =\underset{R_{inl}}\iint{f(x,y) d x d y},
\end{aligned}
\end{equation}
where $f(x,y)$ is the probability density function of $\boldsymbol{p}_i$ over $R_{srch}$. The form of the function $Q(\vartheta)$ depends only on the distribution of $\boldsymbol{p}_i$, while its value is determined by the actual value of $\boldsymbol{p}_i$. Clearly, under the null hypothesis, the probability of observing a localization result with better geometric quality, i.e., a smaller $\theta_i$, is lower.

According to (\ref{eq:geo_err}), the geometric rigidity of the i-th pose ${\hat{\boldsymbol{p}}}_i$ in the set $\mathcal{S}_G$ under camera position $\boldsymbol{p}_c$ is defined as:
\begin{equation}
\begin{aligned}
\alpha_{\boldsymbol{p}_c}(\mathcal{S}_G,i) := {\operatorname{\theta}_i({\hat{\boldsymbol{p}}}_i)}.
\label{qe:alpha_i}
\end{aligned}
\end{equation}
In a CVL process, the geometric rigidity of the entire pose set $\mathcal{S}_G$ is more important than that of any individual sliced pose. We use the maximum normal form function, similar to previous studies \cite{moisan2004probabilistic}, \cite{moisan2012automatic}, to describe the geometric measure $\alpha_{\boldsymbol{p}_c}\left(\mathcal{S}_G\right)$ of the $\mathcal{S}_G$ and identify gross errors:
\begin{equation}
{\alpha_{\boldsymbol{p}_c}\left(\mathcal{S}_G\right)} := {\underset{i}{\operatorname{\max}}\{\alpha_{\boldsymbol{p}_c}\left(\mathcal{S}_G,i\right)\}}.
\label{qe:alpha_set}
\end{equation}

According to the definition mentioned above, our goal can be formally stated as: for the 3-DoF pose set $\mathcal{S}_G$ of the sliced images, the task is to determine a camera position ${\tilde{\boldsymbol{p}}}_c$ that minimizes the geometric rigidity measure $\alpha_{\boldsymbol{p}_c}(\mathcal{S}_G)$:
\begin{equation}
\begin{aligned}
{\tilde{p}}_c=\underset{\boldsymbol{p}_c}{\operatorname{argmin}}\enspace\alpha_{\boldsymbol{p}_c}(\mathcal{S}_G).
\end{aligned}
\end{equation}

\vspace{10pt}
\noindent
\textit{\textbf{Definition 1:} For a sliced images pose set $\mathcal{S}_G$, it is $\alpha${-}rigid if there exists a point $\boldsymbol{p}_c$ such that the corresponding geometric measure satisfies $\alpha_{p_c}\left(\mathcal{S}_G\right)\le\alpha$.}
\vspace{10pt}

Theoretically, any point in the reference image could represent the ground camera's position. However, traversing all possible points is impractical. Therefore, a common approach is to approximate the optimal value through limited sampling. In the case of slice images, a camera position can be estimated from a sample formed by two sliced poses.

Under the null hypothesis $\mathcal{H}_0$, according to Definition 1, for any $\boldsymbol{p}_c$ determined by two 3-DoF poses which from $\mathcal{S}_G$, if $\mathcal{S}_G$ is $\alpha${-}rigid, then it satisfies:
\begin{equation}
\begin{aligned}
\operatorname{Prob}\left(\alpha_{\boldsymbol{p}_c}\left(\mathcal{S}_G\right)\le\alpha|\mathcal{H}_0\right)\le{Q\left(\alpha\right)}^{n-2}.
\end{aligned}
\label{eq:prob_Qn}
\end{equation}
To proof (\ref{eq:prob_Qn}), for a pose ${\hat{\boldsymbol{p}}}_i=(\boldsymbol{p}_i,\phi_i)$ from set $\mathcal{S}_G$, if $\hat{\boldsymbol{p}_i}$ is used to determine the position $\boldsymbol{p}_i$ of camera, then the angle $\theta_i(\boldsymbol{p}_i)=0$, therefor:
\begin{equation}
\begin{aligned}
\operatorname{Prob}\left(\alpha_{\boldsymbol{p}_c}\left(\mathcal{S}_G,i\right)\le\alpha|\mathcal{H}_0\right)=1.
\end{aligned}
\end{equation}
And if ${\hat{\boldsymbol{p}}}_i$ is not used to compute $\boldsymbol{p}_c$, since $\mathcal{S}_G$ is $\alpha${-}rigid, there satisfies $\alpha_{\boldsymbol{p}_c}(\mathcal{S}_G,i)\le\alpha_{\boldsymbol{p}_c}(\mathcal{S}_G)\le\alpha$. Since the poses ${\hat{\boldsymbol{p}}}_i$ in $\mathcal{S}_G$ are estimated in the same way and each ${\hat{\boldsymbol{p}}}_i$ estimation is conducted independently, all poses are mutually independent and follow the same distribution, it holds:
\begin{equation}
\begin{aligned}
\operatorname{Prob}\left(\alpha_{\boldsymbol{p}_c}(\mathcal{S}_G)\le\alpha|\mathcal{H}_0\right)&=\prod_{i}\operatorname{Prob}\left(\alpha_{\boldsymbol{p}_c}(\mathcal{S}_G,i)\le\alpha|\mathcal{H}_0\right)\\
&=\operatorname{Prob}{\left(\theta\le\alpha\middle|\mathcal{H}_0\right)}^{n-2}\\
&=Q(\alpha)^{n-2}
\end{aligned}
\label{eq:prob_alpah}
\end{equation}

\vspace{-10pt}
\subsection{Null Hypothesis}
In most a-contrario methods, the null hypothesis $\mathcal{H}_0$ is conducted as the independent uniform distribution, following human visual behavior and cognitive habits \cite{desolneux2000meaningful}. Because without valid information, decisions are made entirely at random. However, it is arguable to assume that the deep learning models with extrapolation capabilities would also make completely random decisions. Therefore, in this paper, it is necessary to explore the $\mathcal{H}_0$ of the sliced image’s localization error. Research \cite{desolneux2016contrario} suggests that the null hypothesis should be modified based on the observed data. Inspired by this, we use random simulations to generate the observed data and build a suitable background model for the defined CVL pipeline.

\begin{figure}[t!] 
    \centering
    \includegraphics[width=1.0\linewidth]{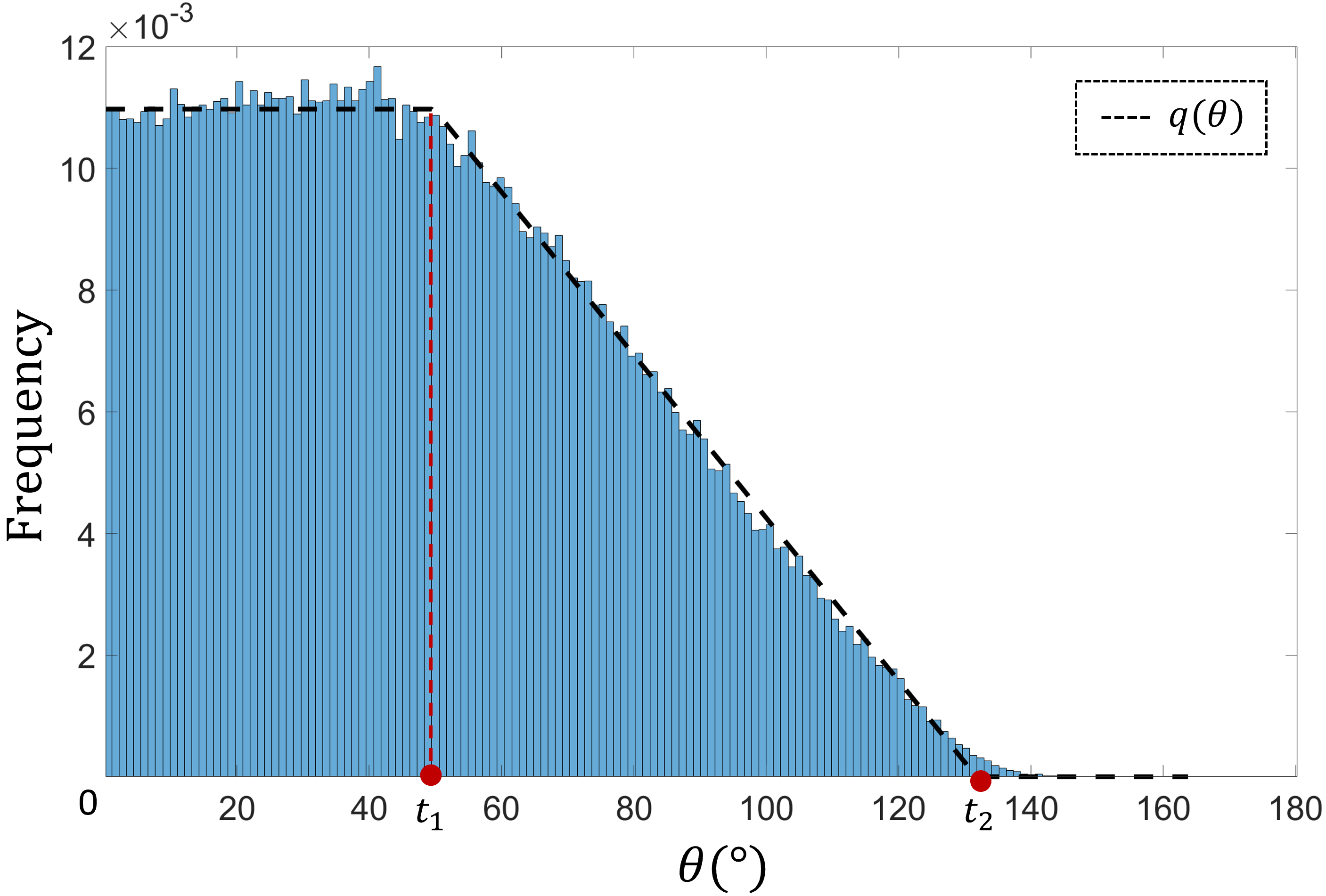}   
    \caption{Distribution of the geometric error $\theta_i$ of slised image’s psoe. On the interval $\left[0,t_1\right)$, $\theta_i$ follows a uniform distribution; on $\left[t_1,t_2\right)$, it follows a linear distribution.}
    \label{fig:fig8}
\vspace{-0.4cm}
\end{figure}

As shown in Fig.~\ref{fig:fig8}, we obtained about $2.6\times{10}^5$ naive poses and plotted the distribution of $\theta_i$, according to (\ref{eq:geo_err}). We found that within a certain range, the frequency decreases linearly as $\theta_i$ increases. Based on the simulations, we define the function as a constant function in the interval $[0,\ t_1)$, a linear function in the interval $[t_1,t_2)$, and zero for $[t_2,\infty)$, as shown in the following:
\begin{equation}
\begin{aligned}
q\left(\theta\right)=
\begin{cases}[r]
C,&0\ \le\theta<t_1 \\ 
A*\theta+B,&t_1\le\theta<t_2 \\ 
0,&t_2\le\theta\end{cases}
\end{aligned}
\end{equation}
where $t_1,t_2,A,B,C$ are parameters. The expression and parameters of the $q\left(\theta\right)$ are independent of the localization data and the CCVPE model parameters; they depend solely on the localization task itself and only need to be determined once. In this paper, we manually set $t_1=50^{\circ}$, $t_2=132^{\circ}$, $A=-6.7\times{10}^{-5}$, $B=8.8\times{10}^{-4}$ according to the simulation. The probability distribution function of $\theta_i$ is defined as:
\begin{equation}
\begin{aligned}
Q\left(\theta\right)=\frac{1}{K}\ast\int_{0}^{\theta}q\left(x\right)dx,\ K=\int_{0}^{+\infty}q\left(x\right)dx.
\end{aligned}
\end{equation}

\vspace{-10pt}
\subsection{Measuring Meaningfulness}
In this section, we employ the geometric rigidity metric and the number of samples to evaluate the meaningfulness of a cross-view localization. The goal is to quantify the "meaning" of pose estimation using a systematic approach. If an estimation is deemed meaningful, the corresponding camera pose is considered valid; otherwise, it should be discarded. We quantify this rigidity metric using the number of false alarms (NFA). NFA represents the expected number of occurrences of event E under the null hypothesis $\mathcal{H}_0$:
\begin{equation}
\begin{aligned}
\text{NFA}=N_{sample}\cdot \operatorname{Prob}\left(E\middle|\mathcal{H}_0\right),
\end{aligned}
\label{eq:nfa_dif}
\end{equation}
$N_{sample}$ denotes the number of samples, $\operatorname{Prob}\left(E\middle|\mathcal{H}_0\right)$ is the occurrence probability of event E. For an event, if the NFA exceeds 1 under $\mathcal{H}_0$, it means the event can easily arise from the background model, making it a meaningless random occurrence not worth further attention.

Here, we say an event E is $\varepsilon$-meaningful if $\operatorname{NFA}(E)\le\varepsilon$. In the Slice-Loc method, if the NFA of a pose set $\mathcal{S}_G$ is less than 1 (which is  $\mathcal{S}_G$ is 1-meaningful), it is a usable case, as it indicates an ordered geometric structure between the sliced image’s poses. Usually, the set $\mathcal{S}_G$ may contain gross errors, and multiple subset samplings and geometric measure evaluations are necessary to identify the inlier set. Referring to \cite{moisan2004probabilistic}, \cite{wan2019contrario}, we use the following approach to determine whether a subset is meaningful.

For the set $\mathcal{S}_G$, a subset $\mathcal{S}_G^{\left(k\right)}$ with $k$ elements is $\varepsilon$-meaningful if $\mathcal{S}_G^{\left(k\right)}$ is $\alpha$-rigid and satisfies:
\begin{equation}
\begin{aligned}
\varepsilon(\alpha,n,k){\colon=}\left(n-2\right)\cdot\left(\begin{matrix}n\\k\\\end{matrix}\right)\cdot\left(\begin{matrix}k\\2\\\end{matrix}\right)\cdot{Q(\alpha)}^{k-2}\le\varepsilon.
\end{aligned}
\label{eq:nfa_est}
\end{equation}
\textit{\textbf{Proof of}} (\ref{eq:nfa_est}): In (\ref{eq:nfa_est}), $(n-2)$ is the number of samplings for $k$, $\left(\begin{matrix}n\\k\\\end{matrix}\right)$ is a combination number for sampling subset $\mathcal{S}_G^{\left(k\right)}$, and $\left(\begin{matrix}k\\2\\\end{matrix}\right)$ is the number of times to sample camera point $\boldsymbol{p}_c$ from $\mathcal{S}_G^{\left(k\right)}$. In practice, not all 3-DoF pose pairs $({\hat{\boldsymbol{p}}}_i,{\hat{\boldsymbol{p}}}_j)\in\mathcal{S}_G\times\mathcal{S}_G$ can be used to estimate point $\boldsymbol{p}_c$ (since some pairs may not intersect), so:
\begin{equation}
\begin{aligned}
N_{sample}\le\left(n-2\right)\cdot\left(\begin{matrix}n\\k\\\end{matrix}\right)\cdot\left(\begin{matrix}k\\2\\\end{matrix}\right).
\end{aligned}
\label{eq:n_sam}
\end{equation}
Combining (\ref{eq:prob_alpah}), (\ref{eq:nfa_dif}) and (\ref{eq:n_sam}):
\begin{equation}
\begin{aligned}
\operatorname{NFA}\left(\alpha_{\boldsymbol{p}_c}\left(\mathcal{S}_k^\prime\right)\le\alpha\middle|\mathcal{H}_0\right)&=N_{sample}\cdot Prob\left(E\middle|\mathcal{H}_0\right)\\
&\le\left(n-2\right)\cdot\left(\begin{matrix}n\\k\\\end{matrix}\right)\cdot\left(\begin{matrix}k\\2\\\end{matrix}\right)Q(\alpha)^{n-2}\\
&=\varepsilon(\alpha,n,k).
\end{aligned}
\end{equation}

\vspace{-10pt}
\subsection{Algorithm}
The subset of $\mathcal{S}_G$ with the smallest $\varepsilon\left(\alpha,n,k\right)$ is regarded as the optimal 3-DoF pose set, denoted as ${\bar{\mathcal{S}}}_G^\prime$. Under the null hypothesis $\mathcal{H}_0$, the set ${\bar{\mathcal{S}}}_G^\prime$ has the smallest expected residual error. The remaining parts of $\mathcal{S}_G$, excluding ${\bar{\mathcal{S}}}_G^\prime$, are classified as outliers. Only the inliers are used to polish the pose of the ground camera. We use the same approach as in \cite{moisan2004probabilistic}, \cite{wan2019contrario} to identify $\mathcal{S}_G^\prime$, which significantly enhances the speed by ranking the residual errors.

Given a camera position $\boldsymbol{p}_c$, each pose ${\hat{\boldsymbol{p}}}_i$ in $\mathcal{S}_G$ is sorted in ascending order according to its geometric rigidity $\alpha_{\boldsymbol{p}_c}\left(\mathcal{S}_G,i\right)$. And the subset $\mathcal{S}_k^\prime$ formed by the first k poses is the subset with the highest geometric rigidity among all possible k-sized subsets of $\mathcal{S}_G$:
\begin{equation}
\begin{aligned}
\alpha_{\boldsymbol{p}_c}(\mathcal{S}_k^\prime)\le
\alpha_{\boldsymbol{p}_c}(\mathcal{S}_k),
\forall\mathcal{S}_k\in
\{\mathcal{S}_k\subseteq\mathcal{S}_G,{|\mathcal{S}}_k|=k\},
\end{aligned}
\end{equation}
therefor:
\begin{equation}
\begin{aligned}
\varepsilon(\alpha_{\boldsymbol{p}_c}(\mathcal{S}_k^\prime),n,k)
\le\varepsilon(\alpha_{\boldsymbol{p}_c}(\mathcal{S}_k),n,k).
\end{aligned}
\end{equation}

By traversing all possible k from 3 to n, the optimal subset corresponding to camera $\boldsymbol{p}_c$ is obtained. The workflow is described in \textbf{Algorithm 1}:
\begin{algorithm}[H]
\caption{get the optimal subset $\bar{\mathcal{S}}_G(\boldsymbol{p}_c)$.}\label{alg:alg1}
\begin{algorithmic}
\STATE \textbf{Input:} pose set $\mathcal{S}_G$, camera point $\boldsymbol{p}_c$.
\STATE \textbf{Output:} optimal subset ${\bar{\mathcal{S}}}_G(\boldsymbol{p}_c)$, reliability indicator scalar $\lg{\bar{\varepsilon}}(\boldsymbol{p}_c)$.
\STATE \textbf{Steps:}
\STATE \hspace{0.4cm}compute the $\alpha_{\boldsymbol{p}_c}(\mathcal{S}_G,i)$ for every ${\hat{\boldsymbol{p}}}_i$ according to (\ref{qe:alpha_i});
\STATE \hspace{0.4cm}sort $\mathcal{S}_G$ according to $\alpha_{\boldsymbol{p}_c}(\mathcal{S}_G,i)$s; 
\STATE \hspace{0.4cm}$\lg{\bar{\varepsilon}}(\boldsymbol{p}_c)\gets\infty $
\STATE \hspace{0.4cm}\textbf{for} $k=3,...,n$ \textbf{do}
\STATE \hspace{0.8cm}construct $\mathcal{S}_k^\prime$ with the first $k$ pose;
\STATE \hspace{0.8cm}compute $\lg{\varepsilon(\alpha,n,k)}$ according to (\ref{eq:nfa_est});
\STATE \hspace{0.8cm}\textbf{if} $\lg{\varepsilon(\alpha,n,k)}<\lg{\bar{\varepsilon}}(\boldsymbol{p}_c)$ \textbf{then}
\STATE \hspace{1.2cm}$\lg{\bar{\varepsilon}}(\boldsymbol{p}_c)\gets \lg{\varepsilon(\alpha,n,k)};$
\STATE \hspace{1.2cm}${\bar{\mathcal{S}}}_G(\boldsymbol{p}_c) \gets \mathcal{S}_k^\prime;$
\STATE \hspace{0.8cm}\textbf{end if}
\STATE \hspace{0.4cm}\textbf{end for}
\STATE \hspace{0.4cm}\textbf{return} $\bar{\mathcal{S}}_G(\boldsymbol{p}_c), \lg{\bar{\varepsilon}}(\boldsymbol{p}_c) $
\end{algorithmic}
\end{algorithm}
The number of sliced images derived from a panoramic image is small (set to 12 in the experiment of this paper), and two 3-DoF poses are enough to determine the camera position $\boldsymbol{p}_c$. Therefore, the number of $\boldsymbol{p}_c$ samples is $\left(\begin{matrix}n\\2\\\end{matrix}\right)$, and the order of magnitude is $\text{O}(n^2)$. We combine all pose pairs in $\mathcal{S}_G$ to determine the corresponding camera position $\boldsymbol{p}_c$, then select the optimal subset ${\bar{\mathcal{S}}}_G(\boldsymbol{p}_c)$ and $\lg{\bar{\varepsilon}}(\boldsymbol{p}_c)$ for each camera position. The set ${\bar{\mathcal{S}}}_G(\boldsymbol{p}_c)$ with the minimum $\lg{\bar{\varepsilon}}(\boldsymbol{p}_c)$ is considered the inlier set, which is then used to refine the ground camera's location.

The whole algorithm named OSA-CVL works as shown in \textbf{Algorithm 2}:
\begin{algorithm}[H]
\caption{Optimized Sampling Algorithm for Cross-view localization.}\label{alg:alg2}
\begin{algorithmic}
\STATE \textbf{Input:} pose set $\mathcal{S}_G$, meaningfulness threshold $\tau=0$.
\STATE \textbf{Output:} optimal subset ${\bar{\mathcal{S}}}_G^\prime$, ground camera location ${\bar{\boldsymbol{p}}}_c$, reliability indicator $\lg{\bar{\varepsilon}}$.
\STATE \textbf{Steps:}
\STATE \hspace{0.4cm}${\bar{\mathcal{S}}}_G^\prime\gets\emptyset$, $\lg{\bar{\varepsilon}}\gets\infty$, ${\bar{\boldsymbol{p}}}_c\gets None$;
\STATE \hspace{0.4cm}construct pose pair set $\mathcal{P}_{r}=\{({\hat{\boldsymbol{p}}}_i,{\hat{\boldsymbol{p}}}_j)\}$;
\STATE \hspace{0.4cm}\textbf{for each} pose-pair $({\hat{\boldsymbol{p}}}_i,{\hat{\boldsymbol{p}}}_j)\in\mathcal{P}_{r}$ \textbf{do}
\STATE \hspace{0.8cm}$\boldsymbol{p}_c \gets \underset{\boldsymbol{p}_c}{\operatorname{argmin}}\left(\theta_i(\hat{\boldsymbol{p}}_i)+\theta_i(\hat{\boldsymbol{p}}_j\right))$ according to (\ref{eq:came_loc});
\STATE \hspace{0.8cm}process \textbf{Algorithm 1} to get ${\bar{\mathcal{S}}}_G(\boldsymbol{p}_c)$ and $\lg{\bar{\varepsilon}}(\boldsymbol{p}_c)$;
\STATE \hspace{0.8cm}\textbf{if} $\lg{\bar{\varepsilon}} < \lg{\bar{\varepsilon}}(\boldsymbol{p}_c)$  \textbf{then}
\STATE \hspace{1.2cm}$\lg{\bar{\varepsilon}}\gets\lg{\bar{\varepsilon}}(\boldsymbol{p}_c);$
\STATE \hspace{1.2cm}${\bar{\mathcal{S}}}_G^\prime \gets{\bar{\mathcal{S}}}_G(\boldsymbol{p}_c);$
\STATE \hspace{0.8cm}\textbf{end if}
\STATE \hspace{0.4cm}\textbf{end for each}
\STATE \hspace{0.4cm}\textbf{if} $\lg{\bar{\varepsilon}} < \tau$  \textbf{then}
\STATE \hspace{0.8cm}// The localization results are available;
\STATE \hspace{0.8cm}refine location ${\bar{\boldsymbol{p}p}}_c$ from ${\bar{\mathcal{S}}}_G^\prime$ according to (\ref{eq:came_loc});
\STATE \hspace{0.4cm}\textbf{else}
\STATE \hspace{0.8cm}// The localization results are not available;
\STATE \hspace{0.8cm}${\bar{\boldsymbol{p}}}_c\gets None$;
\STATE \hspace{0.4cm}\textbf{end if}
\STATE \hspace{0.4cm}\textbf{return} ${\bar{\mathcal{S}}}_G^\prime$, ${\bar{\boldsymbol{p}}}_c$, $\lg{\bar{\varepsilon}}$;
\end{algorithmic}
\end{algorithm}

\section{Comparison Experiment}
\label{experiment}
In this section, we present the comparison results of CVL experiments. First, we describe the datasets, the evaluation metrics, and the implementation details for the proposed Slice-Loc method. Then, Slice-Loc is compared with state-of-the-art (SOTA) methods.

\vspace{-10pt}
\subsection{Datasets}
Slice-Loc is proposed to solve the cross-view localization between satellite-view images and ground-view panoramic images. Therefore, we train and evaluate our method on two cross-view localization datasets: VIGOR \cite{zhu2021vigor} and DReSS \cite{xia2024cross}.

\textit{1) The VIGOR dataset} \cite{zhu2021vigor} is widely used in Cross-View Geo-localization (CVGL) tasks and contains both ground-level and satellite images from four cities in the United States. The dataset includes GPS tags for both ground and aerial images. Ground panoramic images are oriented so that their centers face true north, while aerial images have a size of 640 × 640 pixels at roughly 0.11 m resolution. A ground panorama is considered a positive match for a satellite image if its camera position falls within 1/4 of the aerial image’s width (i.e., 160 pixels) from the reference image’s center. Consistent with prior work, only these positive ground–aerial pairs are used for training and testing. In our experiments, both ground and aerial images are orientation‐aligned: the panorama’s central vertical line corresponds to the North direction in the aerial image. VIGOR offers two data‐split settings: same-area and cross-area. In the same-area setting, the model is trained on half of the data for each city and tested on the other half. In the cross-area setting, the model is trained on images from New York and Seattle, and tested in San Francisco and Chicago. The same-area setting primarily evaluates the method's generalization within the same geographic region, while the cross-area setting tests the method's ability to generalize across different areas. Because VIGOR does not provide the depth maps required by Slice-Loc, we only evaluate in VIGOR’s cross‐area setting.

\textit{2) The DReSS dataset} \cite{xia2024cross} contains ground and aerial images from eight cities across six continents. For our work, we selected six cities: Sydney, Chicago, Johannesburg, Tokyo, Rio, and London, to provide depth maps for training and testing. Each ground image includes a GPS tag. The original DReSS reference maps have a resolution of approximately 0.597 m and cover an area of roughly $132\times132 m^2$. Since competing localization methods expect aerial images at about 0.11 m resolution, we follow \cite{shi2022beyond} to re-collect reference images for each ground panorama. These new reference images are $1280\times1280$ pixels (centered on the ground camera’s GPS position) at approximately 0.11 m resolution, then cropped to $640\times640$ in testing and training. During training, we add random noise to the reference images, uniformly distributed within $\pm160$ pixels as in VIGOR \cite{zhu2021vigor}, and apply random rotation noise of $\pm45^{\circ}$ to ground images. We evaluate under two splits: same-area and cross-area. In the same-area setting, half of the data from the six cities is used for training, and the other half for testing. In the cross-area setting, all the data from Sydney, Chicago, and Johannesburg are used to test the model, while data from Tokyo, Rio, and London are used for training.

\begin{table*}[!t]
    \renewcommand{\arraystretch}{1.1}
    \caption{Evaluation on DReSS Same-Area and Cross-Area Test Set}
    \centering
    \resizebox{\textwidth}{!}{
    \begin{threeparttable}
    \begin{tabular}{ll|ccccc|ccccc}
    \toprule
 \multirow{3}{*}{Noise}& \multirow{3}{*}{Method}& &\multicolumn{3}{c}{Same-Area}& & &\multicolumn{3}{c}{Cross-Area}&\\
 & & {PoS}& \multicolumn{2}{c}{$\downarrow$ Localization (m)}& \multicolumn{2}{c|}{$\downarrow$ Orientation ($^{\circ}$)}
   & {PoS}& \multicolumn{2}{c}{$\downarrow$ Localization (m)}& \multicolumn{2}{c }{$\downarrow$ Orientation ($^{\circ}$)} \\
 & & {(\%)}& mean& median& mean& median& (\%)& mean& median& mean& median \\
\midrule
 \multirow{6}{*}{$0^{\circ}$}& HC-Net \cite{wang2023fine}& \multirow{3}{*}{100}& 3.25& 1.63& 2.50& 1.64& \multirow{3}{*}{100}& 4.51& 2.51& 3.02& 2.00\\
                             & CCVPE \cite{xia2023convolutional}&              & 3.13& 0.99& 2.40& 1.58&                     & 7.39& 2.34& 3.52& 2.42\\
                             & \textbf{Slice-Loc (ours)}        &              & \textbf{2.10}& \textbf{0.82}& \textbf{1.21}& \textbf{0.76}& & \textbf{3.99}& \textbf{1.52}& \textbf{1.37}& \textbf{0.84}\\
                             \cline{2-12}
                             & HC-Net (F)                & \multirow{3}{*}{80.89}& 3.35& 1.67& 2.56& 1.65& \multirow{3}{*}{65.19}& 4.18& 2.33& 2.93& 1.98\\
                             & CCVPE (F)                 &                       & 1.90& 0.85& 2.05& 1.52&                       & 4.32& 1.62& 3.09& 2.29\\
                             & \textbf{Slice-Loc (F)}    &                       & \textbf{1.04}& \textbf{0.71}& \textbf{1.01}& \textbf{0.72}& & \textbf{1.82}& \textbf{1.09}& \textbf{1.09}& \textbf{0.78}\\
\midrule
  \multirow{6}{*}{$20^{\circ}$}& HC-Net \cite{wang2023fine}& \multirow{3}{*}{100}& 3.30& 1.65& 2.64& 1.73& \multirow{3}{*}{100}& 4.57& 2.52& 3.15& 2.08\\
                             & CCVPE \cite{xia2023convolutional}&              & 3.36& 1.03& 2.65& 1.74&                     & 7.73& 2.39& 3.72& 2.37\\
                             & \textbf{Slice-Loc (ours)}        &              & \textbf{2.26}& \textbf{0.85}& \textbf{1.30}& \textbf{0.78}& & \textbf{4.16}& \textbf{1.58}& \textbf{1.48}& \textbf{0.88}\\
                             \cline{2-12}
                             & HC-Net (F)                & \multirow{3}{*}{79.06}& 3.39& 1.67& 2.69& 1.74& \multirow{3}{*}{65.19}& 4.49& 2.48& 3.00& 2.01\\
                             & CCVPE (F)                 &                       & 1.98& 0.88& 2.24& 1.66&                       & 4.43& 1.67& 3.11& 2.20\\
                             & \textbf{Slice-Loc (F)}    &                       & \textbf{1.06}& \textbf{0.73}& \textbf{1.03}& \textbf{0.72}& & \textbf{1.83}& \textbf{1.11}& \textbf{1.12}& \textbf{0.79}\\
\midrule
  \multirow{6}{*}{$45^{\circ}$}& HC-Net \cite{wang2023fine}& \multirow{3}{*}{100}& 3.30& 1.65& 2.64& 1.73& \multirow{3}{*}{100}& 4.82& 2.68& 3.67& 2.22\\
                             & CCVPE \cite{xia2023convolutional}&              & 3.36& 1.04& 2.74& 1.71&                     & 7.78& 2.41& 4.43& 2.28\\
                             & \textbf{Slice-Loc (ours)}        &              & \textbf{2.28}& \textbf{0.86}& \textbf{1.57}& \textbf{0.83}& & \textbf{4.23}& \textbf{1.59}& \textbf{1.87}& \textbf{0.92}\\
                             \cline{2-12}
                             & HC-Net (F)                & \multirow{3}{*}{78.18}& 3.40& 1.67& 2.69& 1.74& \multirow{3}{*}{62.34}& 4.76& 2.65& 3.50& 2.14\\
                             & CCVPE (F)                 &                       & 1.98& 0.88& 2.28& 1.64&                       & 4.47& 1.67& 3.42& 2.12\\
                             & \textbf{Slice-Loc (F)}    &                       & \textbf{1.07}& \textbf{0.73}& \textbf{1.13}& \textbf{0.78}& & \textbf{1.86}& \textbf{1.10}& \textbf{1.24}& \textbf{0.83}\\
\bottomrule
    \end{tabular}
    \begin{tablenotes}[para, flushleft]
    \scriptsize
    {The mean and median of localization and orientation errors for each method in Same-Area and Cross-Area settings are reported. Cross-Area indicates that the model's training and test data come from different city regions, while Same-Area indicates that the training and test data come from the same city region. 0°, 20°, and 45° represent different levels of random rotation applied to the ground images. The results that filter the failed localization are reported for three methods and labeled as (F), and "PoS" denotes the percentage of selected results (select by $\lg{\bar{\varepsilon}}{<}$0). $\downarrow$ indicates that a lower value for this metric is better. The best in bold.}
    \end{tablenotes}
    \end{threeparttable}}
    \label{tab:results_DReSS}
\vspace{-0.4cm}
\end{table*}

\vspace{-10pt}
\subsection{Evaluation Metrics}
We calculate the mean and median errors of the predicted localization and orientation angle for each method across all test data. For localization, the error is the distance between the predicted and the ground truth positions in meters (m). For orientation, the error is the difference between the predicted and the ground truth rotation angles in degrees ($^{\circ}$). Additionally, we compute the proportion of results whose errors fall below certain thresholds. For localization, these thresholds are set at 1 m, 3 m, 5 m, 8 m, and 10 m, while for orientation, the thresholds are set at $1^{\circ}$, $3^{\circ}$, $5^{\circ}$, $8^{\circ}$, and $10^{\circ}$ respectively. In our localization results, errors exceeding 10 m are treated as negative cases. The recall of negative localizations (RoTN) is then computed as:
\begin{equation}
\begin{aligned}
\text{RoTN}=\frac{\text{TN}}{\text{TN}+\text{FP}},
\end{aligned}
\end{equation}
where TN (true negative) are correctly identified failures, and FP (false positives) are incorrect identifications.

\vspace{-10pt}
\subsection{Implementation Details}
The CCVPE model \cite{xia2023convolutional} is chosen to determine the 3-DoFs of the sliced images, and the hyperparameter settings are consistent with those in the original paper and code \cite{xia2023convolutional}. The number of slices during localization is set to $n=12$, meaning one slice is generated every $30^{\circ}$. The HFoV and VFoV of the sliced image are $90^{\circ}$. The pixel size of each sliced image is $512\times512$. All experiments are conducted on a Linux machine equipped with an RTX 4090 GPU. The Pytorch library is used for deep learning network inference and training, with the AdamW optimizer. The batch size is set to 12, and the initial learning rate is 2e-4, which is halved every three epochs for a total of 12 epochs.

In the experiment, we set the reliability determination threshold $\tau=0$, for the Slice-Loc method, only the localization results that passed the reliability check are considered (i.e., $\lg{\bar{\varepsilon}}<\tau$). For other methods, the results are ranked from highest to lowest based on their computed localization confidence scores, and we select the same number of results as Slice-Loc for comparison.

\vspace{-10pt}
\subsection{Results on DReSS Dataset}
We compare the Slice-Loc method with the SOTA cross-view localization methods, CCVPE \cite{xia2023convolutional} and HC-Net \cite{wang2023fine}, on the DReSS \cite{xia2024cross} dataset. For fairness, we retrained both comparison methods based on the official papers and publicly available code. During training, all models were subjected to a random rotation noise of $\pm45^{\circ}$ on the ground panoramic images. During testing, we computed the location and orientation accuracy under three different rotation noises: $0^{\circ}$, $\pm20^{\circ}$, and $\pm45^{\circ}$.

As shown in Table \ref{tab:results_DReSS}, our method achieves smaller location and orientation errors than the two comparison methods in both the Same-Area and Cross-Area settings, demonstrating the superiority of our method in the CVL task.

\textit{1) Compare with CCVPE method:} CCVPE \cite{xia2023convolutional} is currently the highest-performing feature-based method for CVL. However, despite using the same network model for localization, our method exhibits better generalization. In the Cross-Area setting, Slice-Loc reduces the mean localization error by approximately 58\% compared to CCVPE (from 4.47 m to 1.86 m), demonstrating stronger transferability. Moreover, while CCVPE's mean error increases by about 2.5 m (from 1.98 m to 4.47 m) when transitioning from Same-Area to Cross-Area scenarios, Slice-Loc’s error rises by about 0.8 m (from 1.07 m to 1.86 m), highlighting its enhanced robustness. CCVPE determines location by selecting the maximum index from the distribution heatmap, which works well in the Same-Area setting. However, this approach depends on a single information source and struggles in the Cross-Area setting, where the model encounters unfamiliar scenes and fails to produce a correct heatmap, resulting in a significant drop in accuracy. In contrast, Slice-Loc localizes based on sliced images; the final ground-level camera pose is derived from 12 slice poses, each providing independent information for mutual validation. This redundancy enables Slice-Loc to achieve superior localization performance even in untrained scenes.

\begin{table*}[!t]
    \caption{Selected Results on DReSS Dataset}
    \centering
    \resizebox{\textwidth}{!}{
    \begin{threeparttable}
    \begin{tabular}{l|l|c|ccccc|ccccc}
    \toprule
 \multirow{2}{*}{Setting}& \multirow{2}{*}{Method}& $\uparrow$ RoTN& & \multicolumn{3}{c}{$\uparrow$ Localization (\%)}& & & \multicolumn{3}{c}{$\uparrow$ Orientation (\%)}&  \\
 & & {(\%)}& $<$1m& $<$3m& $<$5m& $<$8m& $<$10m& ${<}1^{\circ}$& ${<}3^{\circ}$& ${<}5^{\circ}$& ${<}8^{\circ}$& ${<}10^{\circ}$ \\
\midrule
 \multirow{3}{*}{Same-Area} 
& HC-Net \cite{wang2023fine}       &          16.90& 28.18& 72.08& 83.61& 90.26& 92.41& 30.90& 72.08& 88.13& 95.11& 96.74\\
& CCVPE \cite{xia2023convolutional}&          63.18& 56.91& 90.99& 94.02& 95.28& 95.92& 32.06& 77.21& 93.28& 98.13& 98.81\\
& \textbf{Slice-Loc (ours)}        & \textbf{90.12}& \textbf{67.92}& \textbf{96.80}& \textbf{98.67}& \textbf{99.17}& \textbf{99.32}& \textbf{60.64}& \textbf{94.62}& \textbf{97.91}& \textbf{99.23}& \textbf{99.56}\\
\midrule
  \multirow{3}{*}{Cross-Area}
& HC-Net \cite{wang2023fine}       &          38.54& 14.16& 55.21& 73.46& 84.27& 87.87& 25.34& 63.86& 82.65& 92.61& 94.98\\
& CCVPE \cite{xia2023convolutional}&          64.71& 29.53& 70.29& 80.39& 85.38& 87.30& 25.44& 65.63& 85.46& 94.64& 96.39\\
& \textbf{Slice-Loc (ours)}        & \textbf{88.51}& \textbf{44.59}& \textbf{88.52}& \textbf{95.00}& \textbf{97.28}& \textbf{97.78}& \textbf{57.62}& \textbf{93.65}& \textbf{97.51}& \textbf{98.96}& \textbf{99.42}\\   
\bottomrule
    \end{tabular}
    \begin{tablenotes}[para, flushleft]    
        \scriptsize           
        {Results after filtering out failed localizations. "RoTN" denotes the recall of the true negative (localization error $>$ 10m). The percentage of localization errors below 1m, 3m, 5m, 8m, and 10m and, the percentage of localization errors below 1°, 3°, 5°, 8°, and 10° are listed respectively. The DReSS dataset is used, with an orientation noise level of 45°.}
    \end{tablenotes}            
    \end{threeparttable}}
    \label{tab:results_DReSS_pert}
\vspace{-0.4cm}
\end{table*}
\textit{2) Compare with HC-Net method:} HC-Net \cite{wang2023fine} is a typical geometric-based method that registers the ground image’s BEV map with a reference image, making it the best-performing cross-view fine localization method [18]. However, our Slice-Loc method achieves better pose estimation accuracy. In the Same-Area setting, Slice-Loc reduces the mean localization error by approximately 68\% (from 3.4 m to 1.07 m) and the mean orientation error by 58\% (from $2.69^{\circ}$ to $1.13^{\circ}$) compared to HC-Net. Moreover, HC-Net is less robust to rotation noise; in the Cross-Area setting, as rotation noise increases from $0^{\circ}$ to $45^{\circ}$, HC-Net’s mean localization error rises by 0.58 m, compared to 0.15 m for CCVPE and 0.04 m for Slice-Loc. HC-Net relies on BEV projection to minimize viewpoint differences, but in the complex scenes of the DReSS dataset, the additional noise from the BEV transformation adversely affects accuracy. In contrast, our method relies solely on visual features for localization, which provides stronger generalization across scenes.

It is worth noting that CCVPE performs well on the DReSS dataset. In the Same-Area setting, its pose estimation accuracy exceeds that of HC-Net, which contrasts with the experimental results reported in \cite{wang2023fine}, particularly for orientation. In \cite{wang2023fine}, CCVPE was trained and tested on the VIGOR dataset, where its mean orientation error ($10.56^{\circ}$) was substantially larger than HC-Net’s $2.12^{\circ}$. However, under the same conditions on DReSS, CCVPE achieves an orientation error of $2.74^{\circ}$, comparable to HC-Net’s $2.64^{\circ}$. We attribute this to DReSS’s greater scene diversity, which enhances CCVPE’s generalization in the Same-Area setting and yields more comprehensive, objective results.

\textit{3) Localization score prediction:} The confidence score effectively predicts localization accuracy. In Table~\ref{tab:results_DReSS}, we show the original pose errors. Both CCVPE and Slice-Loc show significant improvements after filtering out failed localizations, whereas HC-Net exhibits only marginal change. Unlike CCVPE’s feature-similarity score, Slice-Loc’s reliability indicator $\lg{\bar{\varepsilon}}$ is derived via geometric verification and, based on a-contrario theory, provides a rigorous probabilistic measure for identifying erroneous localizations with greater efficiency and accuracy. For example, in Table~\ref{tab:results_DReSS_pert}, which reports the recall of true negatives (localization error ${>}10$~m) for all three methods in the Cross-Area setting, Slice-Loc filters out over 88\% of failed cases, nearly 24\% more than CCVPE.

\textit{4) Error distribution:} Table~\ref{tab:results_DReSS_pert} also shows the proportion of localization and orientation errors within specific thresholds under 45°-level rotation noise. For localization, the thresholds are 1 m, 3 m, 5 m, 8 m, and 10 m; and for orientation, they are 1°, 3°, 5°, 8°, and 10°. After filtering out failed localizations, Slice-Loc achieves over 97\% of location errors below 10 m, roughly 10\% better than CCVPE and HC-Net. For orientation, more than 93.7\% of Slice-Loc predictions fall under 3°, roughly 30\% better than CCVPE and HC-Net. This suggests that multiple sub-images with a limited HFoV yield higher orientation accuracy than a single panoramic image with a full HFoV. Slicing the HFoV effectively narrows the feature extraction range, reducing interference among features from different HFoVs.

\begin{figure*}[t] 
    \centering
    \includegraphics[width=1.0\linewidth]{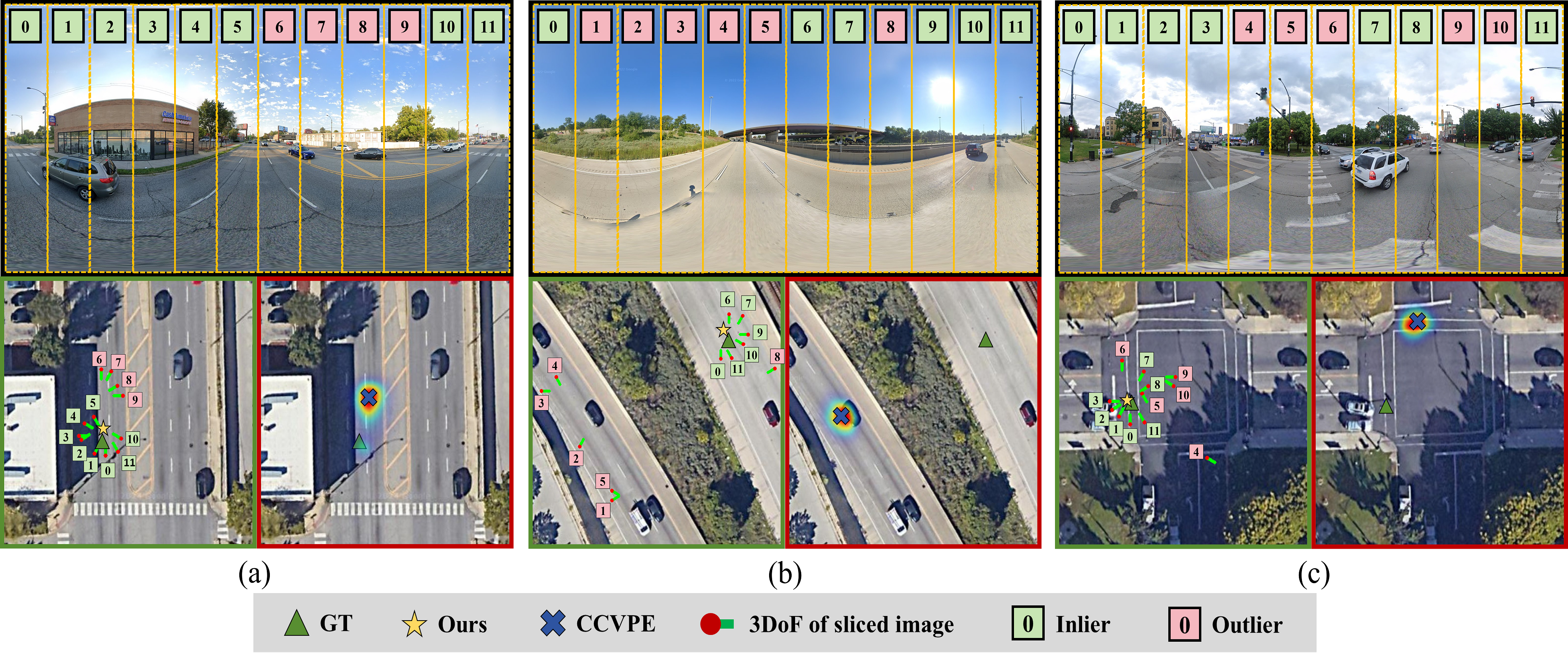}
    \caption{Localization results in scenes with repeated features on the DReSS dataset. For each scene, the upper is the sliced ground image, the bottom left is the localization result of the Slice-Loc method, displaying the 3-DoF rays of the sliced images and the outlier removal results, the bottom right shows the localization heatmap and result of the CCVPE method.}
    \label{fig:fig9}
\vspace{-0.4cm}
\end{figure*}

\textit{5) Qualitative results:} Fig.~\ref{fig:fig9} further examines scenes with symmetric layouts where Slice-Loc outperforms CCVPE. In scene (a), the road runs north-south, so using road features only resolves the east-west position. Consequently, CCVPE exhibits a smaller east-west error but a larger north-south error. In such cases, ground features alone are insufficient, and non-ground features become essential. In the Slice-Loc results, sub-images 6-9 still produce north-south errors, while sub-images 1-4 capture buildings that reduce east-west errors. By discarding erroneous localizations during outlier removal, Slice-Loc achieves superior performance compared to CCVPE. In scene (b), the reference image contains two similar roads, causing CCVPE to localize on the wrong road and incur a large error. Although Slice-Loc also assigns some sub-images to the wrong road, these are identified as outliers due to poor geometric consistency. In contrast, the correct road’s localizations are retained, yielding an accurate final result. Scene (c) depicts a typical urban intersection with four similar intersections (featuring crosswalks and lawns), where CCVPE again mislocalizes. In such scenarios, when visual features alone are insufficient for precise localization, using redundant observations to identify inliers and determine the camera's 3-DoF is a more effective strategy.

\vspace{-10pt}
\subsection{Results on VIGOR  Dataset}
We conducted comparative experiments on the VIGOR Cross-Area set \cite{zhu2021vigor} using several methods: GGCVT \cite{shi2023boosting}, CCVPE \cite{xia2023convolutional}, HC-Net \cite{wang2023fine}, CCVPE-S \cite{xia2024adapting}, and GGCVT-S \cite{xia2024adapting}. The CCVPE-S and GGCVT-S methods improve generalization through unsupervised training. Specifically, we trained the CCVPE and Slice-Loc models on the  DReSS Cross-Area split and tested them on the VIGOR dataset (denoted as *). For Slice-Loc, we report both the original results and the screened results (using $\lg\bar{\varepsilon}<0$), while other methods were evaluated using publicly available models or published results.

Table~\ref{tab:results_VIGOR} presents the detailed localization results. Slice-Loc demonstrates excellent accuracy and generalization. Compared to HC-Net (which achieves SOTA performance on VIGOR), Slice-Loc reduces the mean error by 0.36 meters, with further gains following reliability validation. Moreover, comparing Table \ref{tab:results_DReSS} and Table \ref{tab:results_VIGOR} shows that Slice-Loc achieves significantly better localization accuracy on VIGOR's Cross-Area split than on DReSS's Cross-Area split, confirming that DReSS presents a more challenging environment for CVL.

Additionally, comparisons among CCVPE \cite{xia2023convolutional}, *CCVPE, and CCVPE-S \cite{xia2024adapting} reveal that *CCVPE has a smaller median error and better performance, suggesting that more prosperous and more diverse training data lead to more effective localization models. In summary, *CCVPE, CCVPE-S, and Slice-Loc enhance cross-view fine localization in three distinct ways: *CCVPE uses more complex training data, CCVPE-S improves performance through advanced training strategies, and Slice-Loc employs a superior pose estimation approach. These methods are complementary and can be integrated for practical applications.

\begin{table}[t]
    \caption{Evaluation on VIGOR Cross-Area Test Set}
    \centering
    \resizebox{0.48\textwidth}{!}{
    \begin{threeparttable}
    \begin{tabular}{l|cc|ccc}
    \toprule
    \multirow{2}{*}{Method}& \multicolumn{2}{c|}{$\downarrow$ Localization (m)}& \multicolumn{3}{c}{$\uparrow$ Percentage (\%)}\\
                       &                  mean& median& $<$1m& $<$3m& $<$5m\\
    \midrule
    GGCVT \cite{shi2023boosting}     & 5.16& 1.40& 35.62& 69.07& 74.86\\
    CCVPE \cite{xia2023convolutional}& 4.97& 1.68& 27.37& 70.09& 79.53\\
    HC-Net \cite{wang2023fine}       & 3.36& 1.59& 27.77& 76.23& 86.37\\
    $\dag$CCVPE-S \cite{xia2024adapting}& 3.85& 1.57& 29.36& 74.05& 82.99\\
    $\dag$GGCVT-S \cite{xia2024adapting}& 4.34& 1.32& -& -& -\\
    *CCVPE \cite{xia2023convolutional}  & 6.32& 1.33& 35.95& 77.43& 82.94\\
    *\textbf{Slice-Loc (ours)}          & \textbf{3.00}& \textbf{1.17}& \textbf{41.82}& \textbf{81.67}& \textbf{87.13}\\
    \midrule
    *\textbf{Slice-Loc (F)}             & \textbf{1.65}& \textbf{1.01}& \textbf{49.13}& \textbf{91.78}& \textbf{95.69}\\
    \bottomrule
    \end{tabular}
    \begin{tablenotes}[para, flushleft]
        \scriptsize          
        {The table reports each method’s mean and median localization errors, and the percentages of errors below 1 m, 3 m, and 5 m. "*" indicates that the model is trained on the DReSS Cross-Area train set. "$\dag$" indicates the weakly supervised learning method. "(F)" denotes that the failed localizations have been filtered out.}
    \end{tablenotes}         
    \end{threeparttable}}
    \label{tab:results_VIGOR}
\vspace{-0.4cm}
\end{table}

\section{Discussion}
\label{discussion}
The proposed Slice-Loc method achieved better results for two main reasons: (1) It divides the entire panoramic image into slices to generate redundant observations, then applies a robust estimation process to select the useful information for camera pose estimation. (2) It calculates the NFA of the sliced images’ poses to identify valid geometric patterns in CVL and filter out incorrect camera poses.

To further elucidate Slice-Loc’s localization performance, we present an ablation study, analysis of the NFA, and robust estimation process. Additionally, we offer an analysis of the reliability evaluation.

\vspace{-10pt}
\subsection{Ablation Study}
\label{sec:abl_stu}
We present the ablation study on the DReSS dataset.

\textit{1) Workflow:} The Slice-Loc method comprises three steps: a) slicing and 3-DoF estimation, b) robust estimation, and c) reliability determination. To evaluate each step’s impact, we record the mean and median localization and orientation errors. In Step 1, we use all sliced images’ 3-DoF to estimate the ground camera pose without detecting gross error. Experiments are conducted on the DReSS dataset under the Same-Area setting with 45° rotation noise (see Table \ref{tab:ablation}).

For clarity, Table~\ref{tab:ablation} also presents the results of CCVPE. Comparing CCVPE \cite{xia2023convolutional} with Slice-Loc Step 1 reveals that using multiple slices primarily improves orientation accuracy by effectively narrowing the search space. As the Slice-Loc process progresses, both localization and orientation errors decrease. Specifically, robust estimation reduces the median localization error from 1.57 m to 0.86 m, highlighting a significant improvement after eliminating erroneous 3-DoF estimates. After reliability determination, i.e., removing results with $\lg{\bar{\varepsilon}}>0$, the mean localization error further drops from 2.31 m to 1.07 m, and the percentage of errors over 10 m decreases from 5.61\% to 0.68\%. This shows that reliability determination effectively filters out large errors.

Applying a stricter threshold ($\lg{\bar{\varepsilon}}>-1$) further improves localization performance; however, although the percentage of errors greater than 10 m decreases by about 0.4\%, the PoR drops by roughly 12\%, indicating that most removed errors are under 10 m. This suggests that a threshold of 0 is suitable. In contrast, the improvement in orientation accuracy is minimal, implying that orientation outliers are scarce. Moreover, we found that robust estimation (Step 2) actually worsens orientation accuracy, so we use the mean orientation angle of each sliced image as the camera’s orientation.

\begin{table*}[!ht]
    \caption{Ablation Study on DReSS Same-Area Test Set}
    \centering
    \begin{threeparttable}
    \begin{tabular}{l|l|ccc|ccc|c}
    \toprule
    \multirow{2}{*}{Method}& \multirow{2}{*}{Stage}& \multicolumn{3}{c|}{$\downarrow$ Localization (m)}& \multicolumn{3}{c|}{$\downarrow$ Orientation ($^\circ$)}& \multirow{2}{*}{PoR (\%)}\\
    & & mean& median& $>$10m (\%)& mean& median& $>$10$^\circ$ (\%)& \\
    \midrule
    CCVPE \cite{xia2023convolutional}& -& 3.36& 1.03& 8.64& 2.73& 1.71& 2.37& 100\\
    \midrule
    \multirow{4}{*}{Slice-Loc}       & Step 1& 3.18& 1.57& 6.71& 1.53& 0.85& 1.65& 100\\
                                     & Step 2& 2.31& 0.86& 5.61& 1.57& 0.83& 2.01& 100\\
         & Step 3, $\lg{\bar{\varepsilon}}<0$& 1.07& 0.73& 0.68& 1.14& 0.78& 0.46& 78.18\\
        & Step 3, $\lg{\bar{\varepsilon}}<-1$& 0.89& 0.69& 0.24& 1.04& 0.75& 0.18& 66.70\\
    \bottomrule
    \end{tabular}
    \begin{tablenotes}[para, flushleft]
        \scriptsize 
        {The percentage of localization errors over 10m and orientation errors over 10° are listed. The percentage of reliable localization (PoR) is shown in the last column.}
    \end{tablenotes} 
    \end{threeparttable}
    \label{tab:ablation}
\vspace{-0.4cm}
\end{table*}

\begin{table}[!t]
    \renewcommand{\arraystretch}{1.2}
    \caption{Impact of Slice Number}
    \resizebox{1.0\linewidth}{!}{
    \centering
    \begin{threeparttable}
    \begin{tabular}{c|cccc|cc|c}
    \toprule
    \multirow{2}{*}{$n$}& \multicolumn{2}{c}{$\downarrow$ Localization (m)}& \multicolumn{2}{c|}{$\downarrow$ Orientation ($^\circ$)}& \multicolumn{2}{c|}{$\uparrow$ $\lg{\bar{\varepsilon}}{<}0$ (\%)}& $\downarrow$ Time\\
    & mean& median & mean& median& PoR& $<$10m& (ms)\\
    \midrule
    4& 2.69& 1.15& 1.29& 1.01& 80.00& 93.23& 44\\
    6& 1.66& 0.95& 1.03& 0.81& 72.71& 97.36& 61\\
    8& 1.45& 0.88& 0.98& 0.77& 74.54& 97.99& 85\\
   12& 1.19& 0.84& 0.95& 0.74& 79.37& 99.00& 96\\
   16& 1.18& 0.82& 0.93& 0.72& 81.28& 99.05& 139\\
    \bottomrule
    \end{tabular}
        \label{tab:slice_num}
            \begin{tablenotes}[para, flushleft]
        \scriptsize 
        {"$n$" denotes the number of slices for a ground-level panoramic image. Localization and orientation errors are reported for all data without filtering out failed localizations. "PoR" denotes the proportion of reliable poses, and "$<$10 m" denotes the proportion of errors below 10 m after filtering.}
    \end{tablenotes}
    \end{threeparttable}}
\vspace{-0.4cm}
\end{table}

\textit{2) The number of slices:} The slice number $n$ is a crucial parameter in Slice-Loc, as it determines the number of redundant observations used for localization. In our experiments, we vary $n$ and report the mean and median pose errors (after robust estimation but before reliability determination), the proportion of reliable poses and errors below 10 m after reliability determination, and the localization time. These experiments use the Chicago data from the DReSS Cross-Area setting.

Table~\ref{tab:slice_num} shows that increasing $n$ provides more information for mutual verification, and the defined geometric constraints effectively distinguish inliers from outliers, improving pose estimation accuracy. For instance, when $n$ is reduced from 16 to 4, the RoP remains nearly constant, but the percentage of errors below 10 m drops by 6\%, indicating that at $n=4$, $\lg{\bar{\varepsilon}}>-1$ does not accurately represent the NFA. As $n$ increases from 6 to 16, both the RoP and the proportion of errors under 10 m improve, confirming the effectiveness of $\lg{\bar{\varepsilon}}$.

\begin{table}[!t]
    \renewcommand{\arraystretch}{1.2}
    \caption{Impact of Supervision Type}
    \resizebox{0.98\linewidth}{!}{
    \centering
    \begin{threeparttable}
    \begin{tabular}{lccccccc}
    \toprule
    \multirow{2}{*}{Split}& \multirow{2}{*}{GT}& \multicolumn{3}{c}{$\downarrow$ Localization (m)}& \multicolumn{3}{c}{$\downarrow$ Orientation ($^\circ$)}\\
    & & mean& median& $\uparrow<$5m& mean& median& $\uparrow<$5m\\
    \midrule
    \multirow{2}{*}{Same}& camera& 4.43& 1.24& 78.89& 2.81& 1.72& 90.28\\
                         &  scene& 4.26& 1.36& 80.24& 2.50& 1.50& 92.25\\
    \multirow{2}{*}{Cross}& camera& 7.73& 2.89& 62.15& 3.90& 1.95& 84.94\\
                         &  scene& 6.42& 2.50& 67.87& 2.88& 1.66& 90.11\\
    \bottomrule
    \end{tabular}
                \begin{tablenotes}[para, flushleft]
        \scriptsize 
        {"Camera" indicates that the camera location was used as supervision, while "Scene" indicates that the scene location was used. The error represents the distance between the prediction and the ground truth. The table also lists the percentage of localization errors exceeding 5 m, with a 5° threshold applied for orientation.}
    \end{tablenotes}
    \end{threeparttable}}
    \label{tab:supervision}
\vspace{-0.4cm}
\end{table}

\textit{3) Model training supervision:} Prior cross-view localization studies typically use camera locations as supervision for model optimization. In contrast, our method uses scene locations, which require depth maps and are more challenging to obtain. To evaluate the impact of these two supervision types, we conducted experiments on the DReSS dataset. We trained models using both forms of supervision and evaluated sliced image pose performance in terms of localization and orientation errors. The rotation noise prior was set to 45° during both training and testing.

Table~\ref{tab:supervision} demonstrates that the model trained with scene locations exhibits improved robustness, particularly in the Cross-Area setting. Specifically, the scene-location model shows a lower mean localization error than the camera-location model.

\vspace{-10pt}
\subsection{Distributions of Outliers and Inliers} 
\begin{figure}[t!] 
    \centering
    \includegraphics[width=1.0\linewidth]{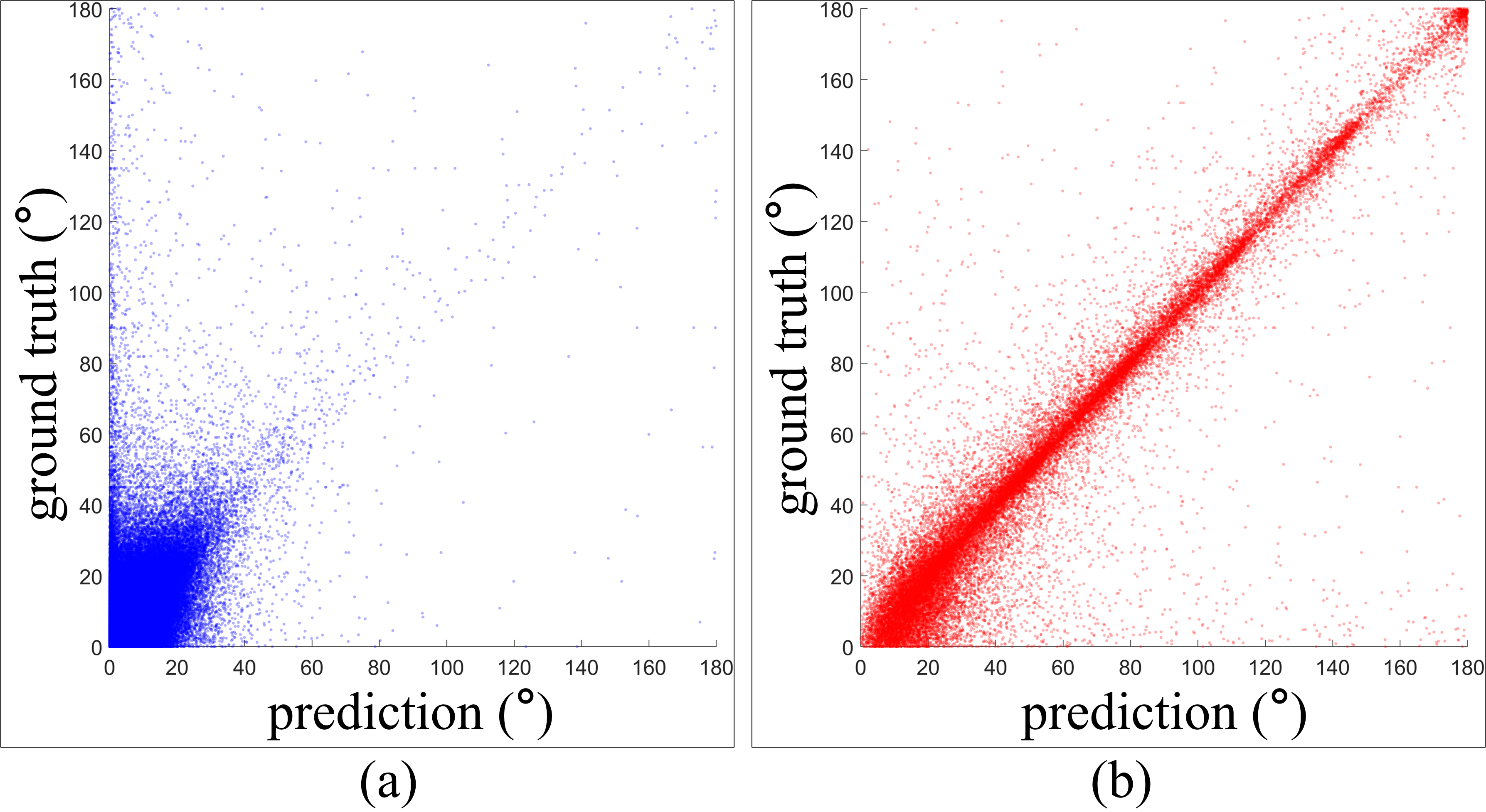}
    \caption{The distribution of sliced poses. (a) is the distribution of inliers, and (b) is the distribution of outliers. The x-axis represents the distance between the predicted camera coordinates and the location of the sub-image, while the y-axis represents the distance between the camera coordinates and the location of the sub-image, with the unit in pixels.}
    \label{fig:fig10}
\vspace{-0.4cm}
\end{figure}

Fig.~\ref{fig:fig10} shows the scatter plot of inlier-outlier identification results, with each point representing the localization of a sliced image. In Fig.~\ref{fig:fig10}(a), the distribution of inlier 3-DoFs is illustrated, with most values falling within the $[0,40]\times[0,40]$ range. In contrast, Fig.~\ref{fig:fig10}(b) reveals that outliers have a more dispersed distribution and exhibit a linear correlation. Even without a fixed threshold, the distinct distributions of inliers and outliers underscore the effectiveness of the defined geometric constraint.

\begin{figure}[t!] 
    \centering
    \includegraphics[width=1.0\linewidth]{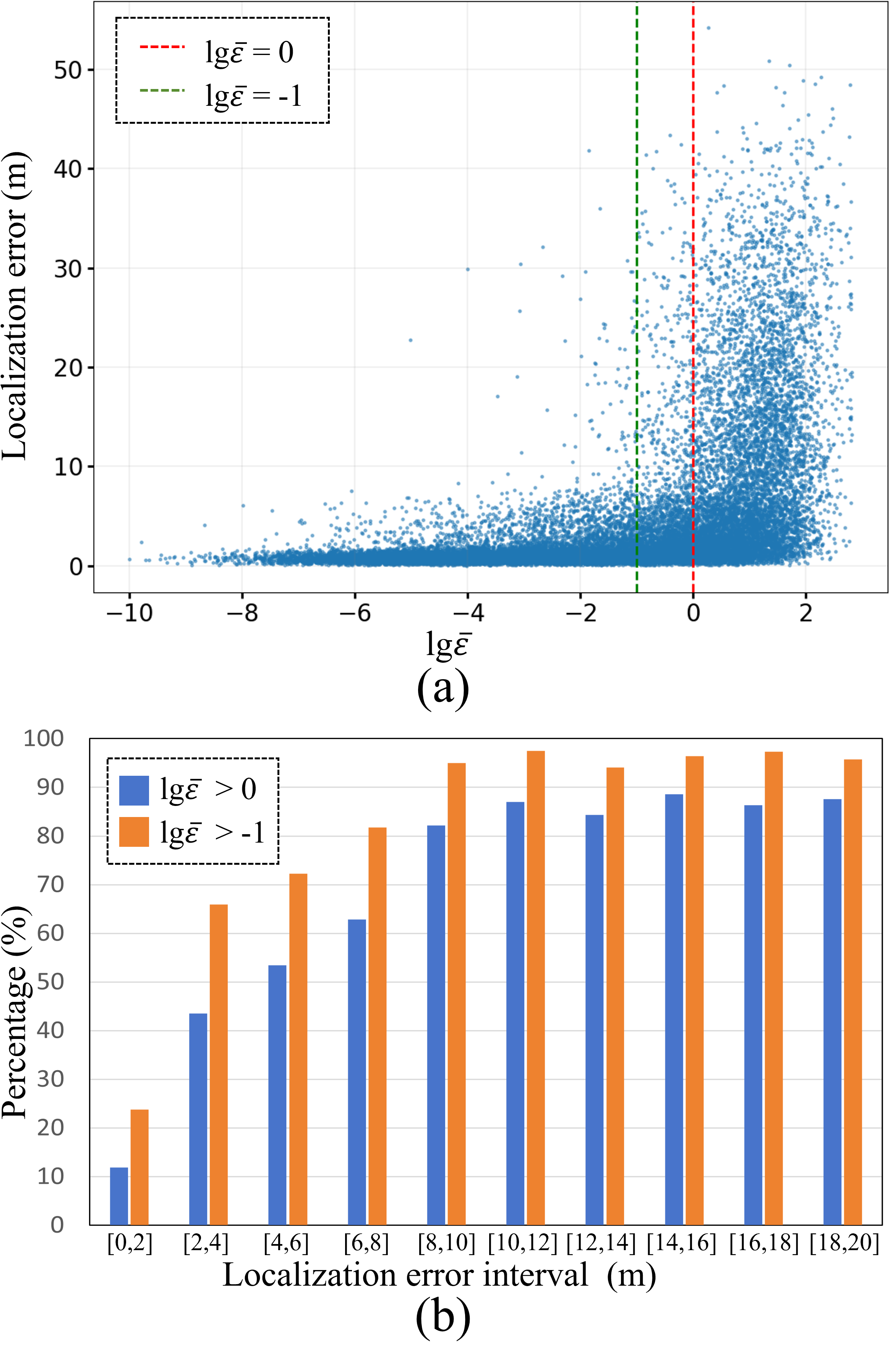}
    \caption{NFA and localization error. (a) Scatter plot of $\lg{\bar{\varepsilon}}$ versus localization error. (b) Percentage of invalid results in each 2 m error interval. The x-axis represents the error intervals (width = 2 m), and the y-axis shows the percentage of invalid results within each interval.}
    \label{fig:fig11}
\vspace{-0.4cm}
\end{figure}

\vspace{-10pt}
\subsection{NFA and Localization Error} 
The Slice-Loc method uses $\lg\bar{\varepsilon}$ to estimate the Number of False Alarms (NFA) for an observed localization. To illustrate the effect of the NFA, Fig.~\ref{fig:fig11}(a) shows a scatter plot of $\lg\bar{\varepsilon}$ versus camera localization error. Additionally, Fig.~\ref{fig:fig11}(b) presents the percentage of the invalid results ($\lg{\bar{\varepsilon}}>0$ and $\lg{\bar{\varepsilon}}>-1$) within each 2-meter interval starting from 0 m.

Fig.~\ref{fig:fig11}(a) reveals that lower $\lg\bar{\varepsilon}$ values correspond to lower localization errors, while higher values are associated with increased errors; most incorrect localizations (errors $>$ 10 m) have $\lg\bar{\varepsilon}>0$. Fig.~\ref{fig:fig11}(b) shows that for small errors, most localizations are considered reliable. As the error increases, the percentage of invalid results also increases, stabilizing beyond 10 m. This suggests small localization errors indicate strong geometric consistency among the CVL tasks. By identifying and filtering out those with poor consistency, the robustness of CVL is effectively improved.

\vspace{-10pt}
\subsection{NFA and Reference Validation} 
A key purpose of estimating NFA is to evaluate the correctness of the reference image provided by the upstream task. Intuitively, if the scene in the ground image matches the reference image, the localization NFA will be small; otherwise, it will be larger. To verify this mechanism, we perform localization using mismatched reference images from two sources: 1) randomly selected and 2) obtained via cross-view retrieval.

\textit{1) Random Selection:} In our experiment, we simulate the null hypothesis $\mathcal{H}_0$ by randomly selecting reference images for each ground query image. Using ground-aerial pairs from the DReSS Same-Area set, we split the data into two halves: one half contains the correct reference image (i.e., the scene visible in the ground panorama), and the other half contains a randomly chosen reference image, ensuring the scene is absent. We set the NFA threshold $\tau$ to 0, such that $\lg\bar{\varepsilon}<\tau$ indicates a correct reference image; otherwise, it is deemed incorrect. We record four types of outcomes: True Positive (TP), False Positive (FP), True Negative (TN), and False Negative (FN), and focus on reporting the negative precision (PoTN), negative recall (RoTN), F1 score (F1), and overall accuracy (Acc).

PoTN indicates the reliability of NFA in correctly identifying negative reference images, and RoTN indicates its accuracy. They are defined as follows:
\begin{equation}
\begin{aligned}
\text{PoTN}=\frac{\text{TN}}{\text{TN}+\text{FN}},
\text{RoTN}=\frac{\text{TN}}{\text{TN}+\text{FP}}.
\end{aligned}
\end{equation}
F1 is the harmonic mean of PoTN and RoTN. F1 and Acc are defined as follows:
\begin{equation}
\begin{aligned}
\text{F1}=2\cdot\frac{\text{PoTN}\times {\text{RoTN}}}{\text{PoTN}+\text{RoTN}},
\text{Acc}=\frac{\text{TN}+\text{TP}}{\text{TN}+\text{FN}+\text{TP}+\text{FP}}.
\end{aligned}
\end{equation}

As shown in Table~\ref{tab:ref_valid1}, benefiting from the null hypothesis design for the CCVPE model, the RoTN reaches around 90\%, demonstrating that the Slice-Loc method can effectively identify mismatched image pairs. The method is more reliable on training data (scenes the model has learned) than test data (scenes the model has not learned).

\begin{figure*}[t!] 
    \centering
    \includegraphics[width=1.0\linewidth]{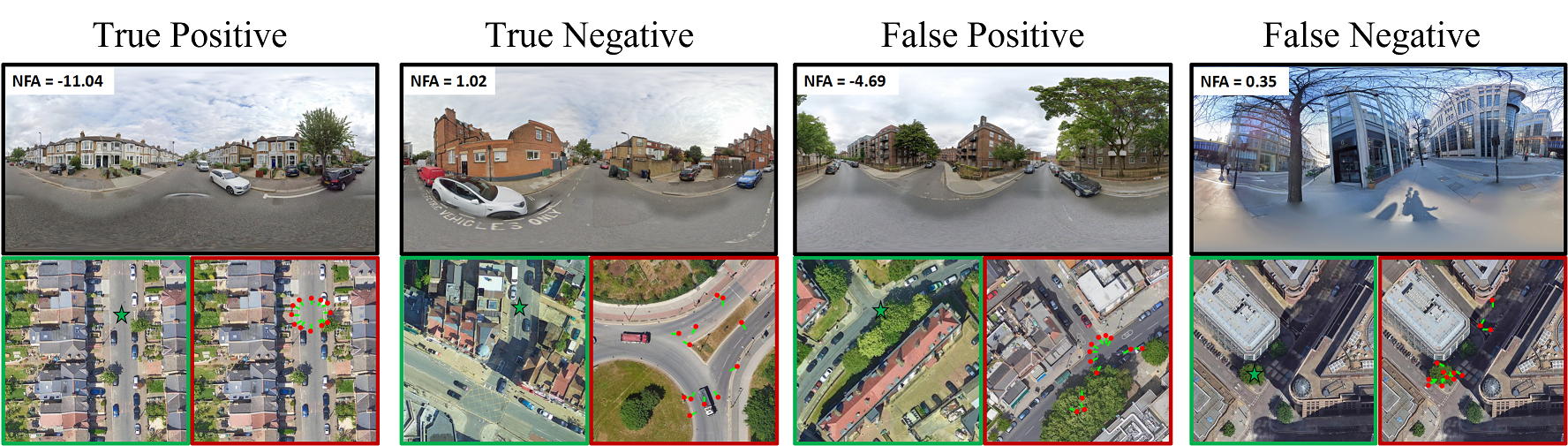}
    \caption{Illustration of the reference validation. The upper is the ground image. The bottom left shows the correct reference image and the ground camera position, while the bottom right indicates the reference image used for positioning, along with the 3-DoFs of each sliced image.}
    \label{fig:fig12}
\vspace{-0.4cm}
\end{figure*}

Fig~\ref{fig:fig12} shows the four validation results. The correct reference map for True Positives yields a clear convergence of the 3-DoF vectors to a single point, demonstrating strong geometric consistency. For True Negatives, an incorrect aerial-view image results in chaotic localization distributions with no clear geometric structure, with an NFA of 1.02, indicating unreliability. In False Positives, although the reference is incorrect, certain visual structures (the road intersection) resemble the correct scene, resulting in relatively good geometric consistency and an NFA of -4.96, which erroneously passes the judgment. For False Negatives, even with correct aerial images, occlusion, and seasonal changes lead to localization in the correct area but with confused orientation, resulting in low geometric consistency and a failed localization. The reliability determination is based on a-contrario principles, by estimating geometric consistency, we can identify visually meaningful structures. In other words, for a given query image, a lower NFA indicates a higher similarity between the reference map and the query scene, as demonstrated by the False Positive result.

\begin{table}[t!]
    \renewcommand{\arraystretch}{1.2}
    \caption{Results of Random Reference Validation}
    \centering
    \begin{threeparttable}
    \begin{tabular}{>{\raggedright\arraybackslash}p{0.15\columnwidth}>{\centering\arraybackslash}p{0.15\columnwidth}>{\centering\arraybackslash}p{0.15\columnwidth}>{\centering\arraybackslash}p{0.15\columnwidth}>{\centering\arraybackslash}p{0.15\columnwidth}}
    \toprule
    & PoTN& RoTN& F1& Acc\\
    \midrule
     Test data& 82.81& 90.53& 86.51& 85.88\\
    Train data& 90.42& 90.62& 90.52& 90.61\\
    \bottomrule
    \end{tabular}
        \begin{tablenotes}[para, flushleft]
        \scriptsize 
        {Test data refers to the results on the test set, and Train data refers to the results on the train set. The unit is \%.}
    \end{tablenotes}
    \end{threeparttable}
    \label{tab:ref_valid1}
\vspace{-0.4cm}
\end{table}

\begin{figure}[t] 
    \centering
    \includegraphics[width=1.0\linewidth]{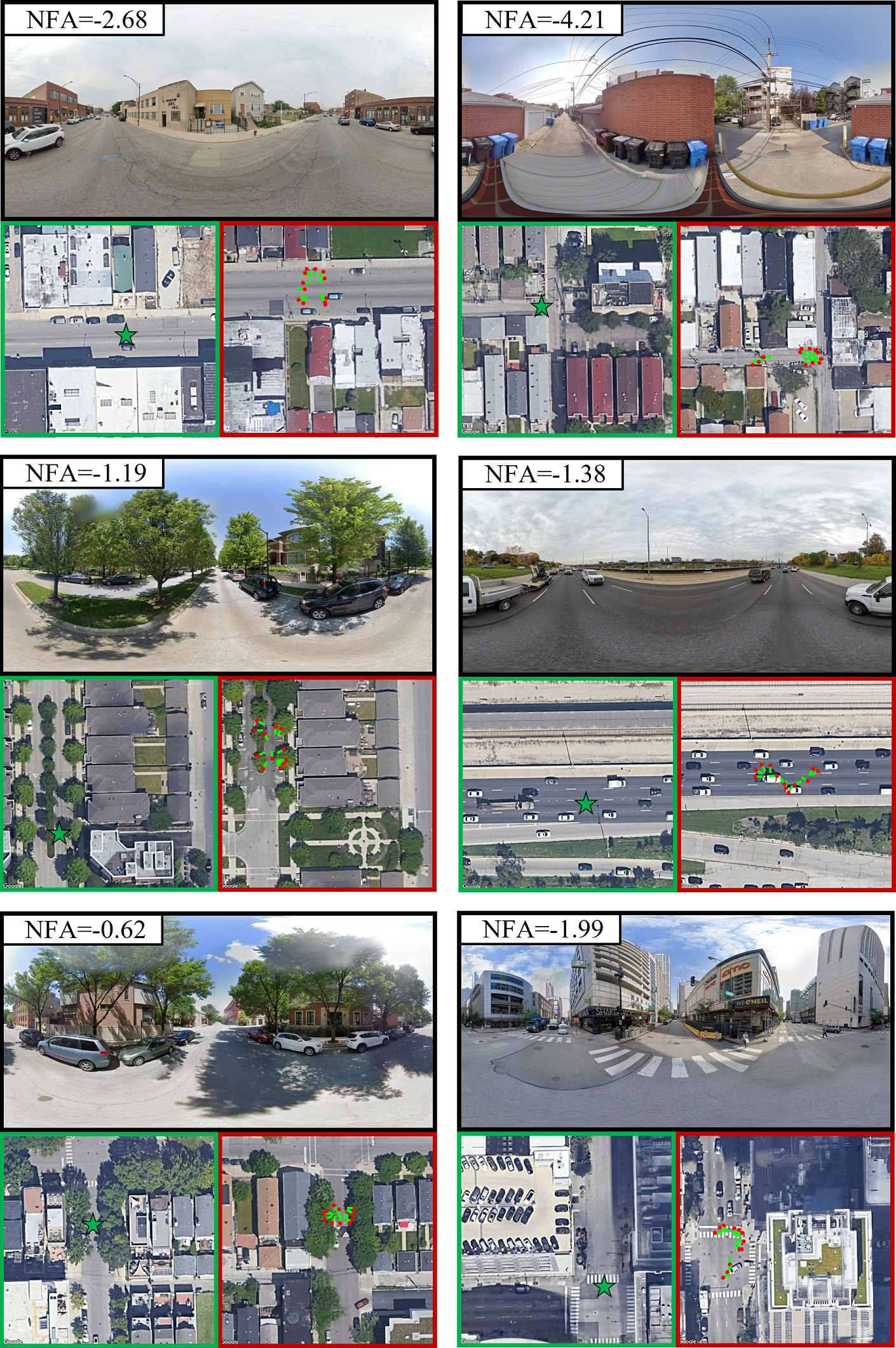}
    \caption{Illustration of the False Positive. In each case, the top shows the ground image; the bottom left shows the correct reference image and the ground camera position; and the bottom right shows the reference provided by Sample4Geo, with the 3-DoFs of each sliced image.}
    \label{fig:fig13}
\vspace{-0.4cm}
\end{figure}

\textit{2) Cross-view retrieval method:} In the CVGL workflow, reference aerial images are retrieved from a database by comparing feature similarities; however, many retrieved references may be incorrect. We conducted experiments on the Chicago subset of the VIGOR dataset using the SOTA retrieval method Sample4Geo \cite{deuser2023sample4geo} to select reference images. We use rough GPS information (1km accuracy) in experiments to identify the wrong reference. We also apply Slice-Loc on the retrieved ground-aerial image pairs to detect the failed retrieved results. In our experiments, in addition to the metrics in Table~\ref{tab:ref_valid1}, we calculate the mean localization error and recalculate the R@1 based on the determination results.

As shown in Table~\ref{tab:ref_valid2}, the retrieval R@1 improves from 61.3\% to 89.82\%, and the average localization error decreases from 461.5 m to 20.4 m due to the filtering of failed retrievals and the fine-grained localization. Initially, using rough GPS data reduces the geo-localization error from 461.5 m to 61 m; however, the RoTN for rough GPS is only 39\%, indicating ineffective identification of negative references. By applying Slice-Loc with 1km-GPS and evaluating geometric consistency, the system’s failure recognition is significantly enhanced, leading to improved accuracy.

However, compared to the results in Table~\ref{tab:ref_valid1}, the RoTN of Slice-Loc in Table~\ref{tab:ref_valid2} declines due to an increased number of FPs (RoTN from 90.53\% to 76.59\%). This is mainly because the negative reference images provided by Sample4Geo exhibit higher geometric similarity to the correct reference images, resulting in inaccurate NFA values, as shown in Fig~\ref{fig:fig13}. Given that Sample4Geo uses high feature similarity reference maps as negative samples during training \cite{deuser2023sample4geo}, we believe the FP results from Slice-Loc can serve as more informative samples. 

\begin{table*}[t!]
    \renewcommand{\arraystretch}{1.2}
    \caption{Results of Retrieved Reference Validation}
    \resizebox{0.99\linewidth}{!}{
    \centering
    \begin{threeparttable}
    \begin{tabular}{p{3.1cm}>{\centering\arraybackslash}p{1.1cm}>{\centering\arraybackslash}p{1.1cm}>{\centering\arraybackslash}p{1.1cm}>{\centering\arraybackslash}p{1.1cm}>{\centering\arraybackslash}p{1.1cm}>{\centering\arraybackslash}p{1.1cm}>{\centering\arraybackslash}p{1.1cm}>{\centering\arraybackslash}p{1.1cm}>{\centering\arraybackslash}p{1.1cm}}
    \toprule
    & TP& FP& TN& FN& PoTN& RoTN& Acc& R@1& Avg (m)\\
    \midrule
    Retrieval& 15641& 9838& 0& 0& -& 0.0& 61.38& 61.38& 461.5\\
    GPS-1km& 15641& 5906& 3932& 0& 100& 39.96& 76.82& 72.59& 61.1\\
    Slice-Loc& 12698& 2303& 7535& 2943& 71.91& 76.59& 79.41& 84.64& 172.1\\
    GPS-1km+Slice-Loc& 12698& 1438& 8400& 2943& 74.05& 85.38& 82.80& 89.82& 20.4\\
    \bottomrule
    \end{tabular}
        \begin{tablenotes}[para, flushleft]
        \scriptsize 
        {The first row (Retrieval) shows the results of Sample4Geo. The second row (GPS-1km) indicates that the failed retrieval results are identified by rough GPS.}
    \end{tablenotes}
    \end{threeparttable}}
    \label{tab:ref_valid2}
\vspace{-0.4cm}
\end{table*}

\section{Conclusion}
\label{conclusion}
In this work, we propose a novel reliability-aware cross-view localization method called Slice-Loc to determine the 3-DoF pose of ground cameras. Slice-Loc estimates the 3-DoF poses of sub-scenes within the sliced HFoV of the query image and estimates the camera pose by merging these sliced images' poses. Based on a-contrario theory, a probabilistic definition under the null hypothesis is proposed for the robust estimation and reliability evaluation algorithm, OSA-CVL. OSA-CVL selects the most meaningful slice pose subset by computing the metric $\bar{\varepsilon}\left(\alpha,n,k\right)$ as an upper bound for the NFA. When $\lg{\bar{\varepsilon}}<0$ is met for the subset, the predicted pose is considered valid for further use; otherwise, the input images should be reexamined to prevent large localization errors.

To achieve pixel-level correspondence, we have established a dataset that offers a more challenging and practical benchmark for cross-view localization, providing dense supervision for training the CVL model. Extensive experiments show that Slice-Loc reduces the mean localization error by 58\% compared to the previous SOTA (from 4.47 m to 1.86 m) in Cross-Area setting, delivering large gains in accuracy and generalization. Moreover, the NFA-based reliability mechanism accurately flags 88\% of the failed localizations, which are prone to large errors. Our ablation studies further reveal that effective cross-view localization requires identifying features that hinder positioning and detecting scenes where localization fails. In our framework, Slice-Loc integrates retrieval-based geo-localization with feature-based fine localization, yielding an interpretable and robust localization pipeline.

By leveraging denser scene-level matching to provide independent and redundant observations, Slice-Loc enhances cross-view pose estimation by detecting meaningful geometry patterns. However, the limited performance of the CCVPE model constrains its ability to generate many redundant observations within a single CVL process, thereby limiting Slice-Loc’s failure detection precision and overall localization performance. Building on the proposed dataset and method, we plan to explore deep learning models capable of implicit cross-view feature matching to achieve reliable and efficient cross-view localization ultimately.

\bibliographystyle{IEEEtran}
\bibliography{reference}

\begin{thebibliography}{10}
\providecommand{\url}[1]{#1}
\csname url@samestyle\endcsname
\providecommand{\newblock}{\relax}
\providecommand{\bibinfo}[2]{#2}
\providecommand{\BIBentrySTDinterwordspacing}{\spaceskip=0pt\relax}
\providecommand{\BIBentryALTinterwordstretchfactor}{4}
\providecommand{\BIBentryALTinterwordspacing}{\spaceskip=\fontdimen2\font plus
\BIBentryALTinterwordstretchfactor\fontdimen3\font minus \fontdimen4\font\relax}
\providecommand{\BIBforeignlanguage}[2]{{%
\expandafter\ifx\csname l@#1\endcsname\relax
\typeout{** WARNING: IEEEtran.bst: No hyphenation pattern has been}%
\typeout{** loaded for the language `#1'. Using the pattern for}%
\typeout{** the default language instead.}%
\else
\language=\csname l@#1\endcsname
\fi
#2}}
\providecommand{\BIBdecl}{\relax}
\BIBdecl

\bibitem{zheng2020university}
Z.~Zheng, Y.~Wei, and Y.~Yang, ``University-1652: A multi-view multi-source benchmark for drone-based geo-localization,'' in \emph{Proceedings of the 28th ACM international conference on Multimedia}, 2020, pp. 1395--1403.

\bibitem{srivastava2019understanding}
S.~Srivastava, J.~E. Vargas-Munoz, and D.~Tuia, ``Understanding urban landuse from the above and ground perspectives: A deep learning, multimodal solution,'' \emph{Remote sensing of environment}, vol. 228, pp. 129--143, 2019.

\bibitem{feng2018urban}
T.~Feng, Q.-T. Truong, D.~T. Nguyen, J.~Y. Koh, L.-F. Yu, A.~Binder, and S.-K. Yeung, ``Urban zoning using higher-order markov random fields on multi-view imagery data,'' in \emph{Proceedings of the European Conference on Computer Vision (ECCV)}, 2018, pp. 614--630.

\bibitem{maddern20171}
W.~Maddern, G.~Pascoe, C.~Linegar, and P.~Newman, ``1 year, 1000 km: The oxford robotcar dataset,'' \emph{The International Journal of Robotics Research}, vol.~36, no.~1, pp. 3--15, 2017.

\bibitem{li2025cross}
H.~Li, F.~Deuser, W.~Yin, X.~Luo, P.~Walther, G.~Mai, W.~Huang, and M.~Werner, ``Cross-view geolocalization and disaster mapping with street-view and vhr satellite imagery: A case study of hurricane ian,'' \emph{ISPRS Journal of Photogrammetry and Remote Sensing}, vol. 220, pp. 841--854, 2025.

\bibitem{ye2024sg}
J.~Ye, Q.~Luo, J.~Yu, H.~Zhong, Z.~Zheng, C.~He, and W.~Li, ``Sg-bev: satellite-guided bev fusion for cross-view semantic segmentation,'' in \emph{Proceedings of the IEEE/CVF Conference on Computer Vision and Pattern Recognition}, 2024, pp. 27\,748--27\,757.

\bibitem{sun2023cross}
Y.~Sun, Y.~Ye, J.~Kang, R.~Fernandez-Beltran, S.~Feng, X.~Li, C.~Luo, P.~Zhang, and A.~Plaza, ``Cross-view object geo-localization in a local region with satellite imagery,'' \emph{IEEE Transactions on Geoscience and Remote Sensing}, vol.~61, pp. 1--16, 2023.

\bibitem{ye2024coarse}
Q.~Ye, J.~Luo, and Y.~Lin, ``A coarse-to-fine visual geo-localization method for gnss-denied uav with oblique-view imagery,'' \emph{ISPRS Journal of Photogrammetry and Remote Sensing}, vol. 212, pp. 306--322, 2024.

\bibitem{zhu2021vigor}
S.~Zhu, T.~Yang, and C.~Chen, ``Vigor: Cross-view image geo-localization beyond one-to-one retrieval,'' in \emph{Proceedings of the IEEE/CVF Conference on Computer Vision and Pattern Recognition}, 2021, pp. 3640--3649.

\bibitem{workman2015wide}
S.~Workman, R.~Souvenir, and N.~Jacobs, ``Wide-area image geolocalization with aerial reference imagery,'' in \emph{Proceedings of the IEEE International Conference on Computer Vision}, 2015, pp. 3961--3969.

\bibitem{liu2019lending}
L.~Liu and H.~Li, ``Lending orientation to neural networks for cross-view geo-localization,'' in \emph{Proceedings of the IEEE/CVF conference on computer vision and pattern recognition}, 2019, pp. 5624--5633.

\bibitem{xia2024cross}
P.~Xia, L.~Yu, Y.~Wan, Q.~Wu, P.~Chen, L.~Zhong, Y.~Yao, D.~Wei, X.~Liu, L.~Ru \emph{et~al.}, ``Cross-view geo-localization with street-view and vhr satellite imagery in decentrality settings,'' \emph{arXiv preprint arXiv:2412.11529}, 2024.

\bibitem{geiger2013vision}
A.~Geiger, P.~Lenz, C.~Stiller, and R.~Urtasun, ``Vision meets robotics: The kitti dataset,'' \emph{The international journal of robotics research}, vol.~32, no.~11, pp. 1231--1237, 2013.

\bibitem{agarwal2020ford}
S.~Agarwal, A.~Vora, G.~Pandey, W.~Williams, H.~Kourous, and J.~McBride, ``Ford multi-av seasonal dataset,'' \emph{The International Journal of Robotics Research}, vol.~39, no.~12, pp. 1367--1376, 2020.

\bibitem{shen2023mccg}
T.~Shen, Y.~Wei, L.~Kang, S.~Wan, and Y.-H. Yang, ``Mccg: A convnext-based multiple-classifier method for cross-view geo-localization,'' \emph{IEEE Transactions on Circuits and Systems for Video Technology}, vol.~34, no.~3, pp. 1456--1468, 2023.

\bibitem{ye2024cross}
J.~Ye, Z.~Lv, W.~Li, J.~Yu, H.~Yang, H.~Zhong, and C.~He, ``Cross-view image geo-localization with panorama-bev co-retrieval network,'' in \emph{European Conference on Computer Vision}.\hskip 1em plus 0.5em minus 0.4em\relax Springer, 2024, pp. 74--90.

\bibitem{deuser2023sample4geo}
F.~Deuser, K.~Habel, and N.~Oswald, ``Sample4geo: Hard negative sampling for cross-view geo-localisation,'' in \emph{Proceedings of the IEEE/CVF International Conference on Computer Vision}, 2023, pp. 16\,847--16\,856.

\bibitem{wang2023fine}
X.~Wang, R.~Xu, Z.~Cui, Z.~Wan, and Y.~Zhang, ``Fine-grained cross-view geo-localization using a correlation-aware homography estimator,'' \emph{Advances in Neural Information Processing Systems}, vol.~36, pp. 5301--5319, 2023.

\bibitem{shi2022geometry}
Y.~Shi, D.~Campbell, X.~Yu, and H.~Li, ``Geometry-guided street-view panorama synthesis from satellite imagery,'' \emph{IEEE Transactions on Pattern Analysis and Machine Intelligence}, vol.~44, no.~12, pp. 10\,009--10\,022, 2022.

\bibitem{piasco2018survey}
N.~Piasco, D.~Sidib{\'e}, C.~Demonceaux, and V.~Gouet-Brunet, ``A survey on visual-based localization: On the benefit of heterogeneous data,'' \emph{Pattern Recognition}, vol.~74, pp. 90--109, 2018.

\bibitem{noh2017large}
H.~Noh, A.~Araujo, J.~Sim, T.~Weyand, and B.~Han, ``Large-scale image retrieval with attentive deep local features,'' in \emph{Proceedings of the IEEE international conference on computer vision}, 2017, pp. 3456--3465.

\bibitem{fervers2023c}
F.~Fervers, S.~Bullinger, C.~Bodensteiner, M.~Arens, and R.~Stiefelhagen, ``C-bev: Contrastive bird's eye view training for cross-view image retrieval and 3-dof pose estimation,'' \emph{arXiv preprint arXiv:2312.08060}, 2023.

\bibitem{fervers2023uncertainty}
{Fervers, Florian and Bullinger, Sebastian and Bodensteiner, Christoph and Arens, Michael and Stiefelhagen, Rainer}, ``Uncertainty-aware vision-based metric cross-view geolocalization,'' in \emph{Proceedings of the IEEE/CVF Conference on Computer Vision and Pattern Recognition}, 2023, pp. 21\,621--21\,631.

\bibitem{shi2022beyond}
Y.~Shi and H.~Li, ``Beyond cross-view image retrieval: Highly accurate vehicle localization using satellite image,'' in \emph{Proceedings of the IEEE/CVF Conference on Computer Vision and Pattern Recognition}, 2022, pp. 17\,010--17\,020.

\bibitem{hays2008im2gps}
J.~Hays and A.~A. Efros, ``Im2gps: estimating geographic information from a single image,'' in \emph{2008 ieee conference on computer vision and pattern recognition}.\hskip 1em plus 0.5em minus 0.4em\relax IEEE, 2008, pp. 1--8.

\bibitem{lin2013cross}
T.-Y. Lin, S.~Belongie, and J.~Hays, ``Cross-view image geolocalization,'' in \emph{Proceedings of the IEEE Conference on Computer Vision and Pattern Recognition}, 2013, pp. 891--898.

\bibitem{fu2017fast}
C.~Fu, C.~Xiang, C.~Wang, and D.~Cai, ``Fast approximate nearest neighbor search with the navigating spreading-out graph,'' \emph{arXiv preprint arXiv:1707.00143}, 2017.

\bibitem{lin2015learning}
T.-Y. Lin, Y.~Cui, S.~Belongie, and J.~Hays, ``Learning deep representations for ground-to-aerial geolocalization,'' in \emph{Proceedings of the IEEE conference on computer vision and pattern recognition}, 2015, pp. 5007--5015.

\bibitem{hu2018cvm}
S.~Hu, M.~Feng, R.~M. Nguyen, and G.~H. Lee, ``Cvm-net: Cross-view matching network for image-based ground-to-aerial geo-localization,'' in \emph{Proceedings of the IEEE Conference on Computer Vision and Pattern Recognition}, 2018, pp. 7258--7267.

\bibitem{shi2020optimal}
Y.~Shi, X.~Yu, L.~Liu, T.~Zhang, and H.~Li, ``Optimal feature transport for cross-view image geo-localization,'' in \emph{Proceedings of the AAAI Conference on Artificial Intelligence}, vol.~34, no.~07, 2020, pp. 11\,990--11\,997.

\bibitem{wang2021each}
T.~Wang, Z.~Zheng, C.~Yan, J.~Zhang, Y.~Sun, B.~Zheng, and Y.~Yang, ``Each part matters: Local patterns facilitate cross-view geo-localization,'' \emph{IEEE Transactions on Circuits and Systems for Video Technology}, vol.~32, no.~2, pp. 867--879, 2021.

\bibitem{vaswani2017attention}
A.~Vaswani, N.~Shazeer, N.~Parmar, J.~Uszkoreit, L.~Jones, A.~N. Gomez, {\L}.~Kaiser, and I.~Polosukhin, ``Attention is all you need,'' \emph{Advances in neural information processing systems}, vol.~30, 2017.

\bibitem{yang2021cross}
H.~Yang, X.~Lu, and Y.~Zhu, ``Cross-view geo-localization with layer-to-layer transformer,'' \emph{Advances in Neural Information Processing Systems}, vol.~34, pp. 29\,009--29\,020, 2021.

\bibitem{zhu2022transgeo}
S.~Zhu, M.~Shah, and C.~Chen, ``Transgeo: Transformer is all you need for cross-view image geo-localization,'' in \emph{Proceedings of the IEEE/CVF Conference on Computer Vision and Pattern Recognition}, 2022, pp. 1162--1171.

\bibitem{wu2024crossviewimagesetgeolocalization}
\BIBentryALTinterwordspacing
Q.~Wu, P.~Xia, L.~Yu, Y.~Liu, M.~Xiong, L.~Zhong, J.~Chen, M.~Yang, Y.~Zhang, and Y.~Wan, ``Cross-view image set geo-localization,'' 2024. [Online]. Available: \url{https://arxiv.org/abs/2412.18852}
\BIBentrySTDinterwordspacing

\bibitem{shi2019spatial}
Y.~Shi, L.~Liu, X.~Yu, and H.~Li, ``Spatial-aware feature aggregation for image based cross-view geo-localization,'' \emph{Advances in Neural Information Processing Systems}, vol.~32, 2019.

\bibitem{li2023multi}
S.~Li, Z.~Tu, Y.~Chen, and T.~Yu, ``Multi-scale attention encoder for street-to-aerial image geo-localization,'' \emph{CAAI Transactions on Intelligence Technology}, vol.~8, no.~1, pp. 166--176, 2023.

\bibitem{regmi2018cross}
K.~Regmi and A.~Borji, ``Cross-view image synthesis using conditional gans,'' in \emph{Proceedings of the IEEE conference on Computer Vision and Pattern Recognition}, 2018, pp. 3501--3510.

\bibitem{tang2019multi}
H.~Tang, D.~Xu, N.~Sebe, Y.~Wang, J.~J. Corso, and Y.~Yan, ``Multi-channel attention selection gan with cascaded semantic guidance for cross-view image translation,'' in \emph{Proceedings of the IEEE/CVF conference on computer vision and pattern recognition}, 2019, pp. 2417--2426.

\bibitem{toker2021coming}
A.~Toker, Q.~Zhou, M.~Maximov, and L.~Leal-Taix{\'e}, ``Coming down to earth: Satellite-to-street view synthesis for geo-localization,'' in \emph{Proceedings of the IEEE/CVF Conference on Computer Vision and Pattern Recognition}, 2021, pp. 6488--6497.

\bibitem{hou2022road}
Y.~Hou, Y.~Yang, J.~Wang, and M.~Fu, ``Road extraction assisted offset regression method in cross-view image-based geo-localization,'' in \emph{2022 IEEE 25th International Conference on Intelligent Transportation Systems (ITSC)}.\hskip 1em plus 0.5em minus 0.4em\relax IEEE, 2022, pp. 2934--2940.

\bibitem{hu2022beyond}
W.~Hu, Y.~Zhang, Y.~Liang, Y.~Yin, A.~Georgescu, A.~Tran, H.~Kruppa, S.-K. Ng, and R.~Zimmermann, ``Beyond geo-localization: Fine-grained orientation of street-view images by cross-view matching with satellite imagery,'' in \emph{Proceedings of the 30th ACM international conference on multimedia}, 2022, pp. 6155--6164.

\bibitem{xia2022visual}
Z.~Xia, O.~Booij, M.~Manfredi, and J.~F. Kooij, ``Visual cross-view metric localization with dense uncertainty estimates,'' in \emph{European Conference on Computer Vision}.\hskip 1em plus 0.5em minus 0.4em\relax Springer, 2022, pp. 90--106.

\bibitem{lentsch2023slicematch}
T.~Lentsch, Z.~Xia, H.~Caesar, and J.~F. Kooij, ``Slicematch: Geometry-guided aggregation for cross-view pose estimation,'' in \emph{Proceedings of the IEEE/CVF Conference on Computer Vision and Pattern Recognition}, 2023, pp. 17\,225--17\,234.

\bibitem{xia2023convolutional}
Z.~Xia, O.~Booij, and J.~F. Kooij, ``Convolutional cross-view pose estimation,'' \emph{IEEE Transactions on Pattern Analysis and Machine Intelligence}, vol.~46, no.~5, pp. 3813--3831, 2023.

\bibitem{shi2023boosting}
Y.~Shi, F.~Wu, A.~Perincherry, A.~Vora, and H.~Li, ``Boosting 3-dof ground-to-satellite camera localization accuracy via geometry-guided cross-view transformer,'' in \emph{Proceedings of the IEEE/CVF International Conference on Computer Vision}, 2023, pp. 21\,516--21\,526.

\bibitem{wang2024view}
S.~Wang, C.~Nguyen, J.~Liu, Y.~Zhang, S.~Muthu, F.~A. Maken, K.~Zhang, and H.~Li, ``View from above: Orthogonal-view aware cross-view localization,'' in \emph{Proceedings of the IEEE/CVF Conference on Computer Vision and Pattern Recognition}, 2024, pp. 14\,843--14\,852.

\bibitem{dosovitskiy2020image}
A.~Dosovitskiy, L.~Beyer, A.~Kolesnikov, D.~Weissenborn, X.~Zhai, T.~Unterthiner, M.~Dehghani, M.~Minderer, G.~Heigold, S.~Gelly \emph{et~al.}, ``An image is worth 16x16 words: Transformers for image recognition at scale,'' \emph{arXiv preprint arXiv:2010.11929}, 2020.

\bibitem{xia2024adapting}
Z.~Xia, Y.~Shi, H.~Li, and J.~FP~Kooij, ``Adapting fine-grained cross-view localization to areas without fine ground truth,'' in \emph{European Conference on Computer Vision}.\hskip 1em plus 0.5em minus 0.4em\relax Springer, 2024, pp. 397--415.

\bibitem{shi2024weakly}
Y.~Shi, H.~Li, A.~Perincherry, and A.~Vora, ``Weakly-supervised camera localization by ground-to-satellite image registration,'' in \emph{European Conference on Computer Vision}.\hskip 1em plus 0.5em minus 0.4em\relax Springer, 2024, pp. 39--57.

\bibitem{miao2024survey}
J.~Miao, K.~Jiang, T.~Wen, Y.~Wang, P.~Jia, B.~Wijaya, X.~Zhao, Q.~Cheng, Z.~Xiao, J.~Huang \emph{et~al.}, ``A survey on monocular re-localization: From the perspective of scene map representation,'' \emph{IEEE Transactions on Intelligent Vehicles}, 2024.

\bibitem{liu2025render}
Q.~Liu, Z.~Liu, Q.~Wu, P.~Xia, and Y.~Wan, ``Render then match: A neural radiance field-based indoor visual localization framework for lighting-varying environments using infrared images,'' \emph{The Photogrammetric Record}, vol.~40, no. 189, p. e70002, 2025.

\bibitem{li2021robust}
J.~Li, Y.~Zhang, and Q.~Hu, ``Robust estimation in robot vision and photogrammetry: a new model and its applications,'' \emph{ISPRS Annals of the Photogrammetry, Remote Sensing and Spatial Information Sciences}, vol.~1, pp. 137--144, 2021.

\bibitem{fischler1981random}
M.~FISCHLER~AND, ``Random sample consensus: a paradigm for model fitting with applications to image analysis and automated cartography,'' \emph{Commun. ACM}, vol.~24, no.~6, pp. 381--395, 1981.

\bibitem{mishkin2015mods}
D.~Mishkin, J.~Matas, and M.~Perdoch, ``Mods: Fast and robust method for two-view matching,'' \emph{Computer vision and image understanding}, vol. 141, pp. 81--93, 2015.

\bibitem{li2023qgore}
J.~Li, P.~Shi, Q.~Hu, and Y.~Zhang, ``Qgore: Quadratic-time guaranteed outlier removal for point cloud registration,'' \emph{IEEE Transactions on Pattern Analysis and Machine Intelligence}, vol.~45, no.~9, pp. 11\,136--11\,151, 2023.

\bibitem{moisan2004probabilistic}
L.~Moisan and B.~Stival, ``A probabilistic criterion to detect rigid point matches between two images and estimate the fundamental matrix,'' \emph{International Journal of Computer Vision}, vol.~57, pp. 201--218, 2004.

\bibitem{moisan2012automatic}
L.~Moisan, P.~Moulon, and P.~Monasse, ``Automatic homographic registration of a pair of images, with a contrario elimination of outliers,'' \emph{Image Processing On Line}, vol.~2, pp. 56--73, 2012.

\bibitem{wan2017p2l}
Y.~Wan and Y.~Zhang, ``The p2l method of mismatch detection for push broom high-resolution satellite images,'' \emph{ISPRS Journal of Photogrammetry and Remote Sensing}, vol. 130, pp. 317--328, 2017.

\bibitem{wang2023satellite}
S.~Wang, Y.~Zhang, A.~Vora, A.~Perincherry, and H.~Li, ``Satellite image based cross-view localization for autonomous vehicle,'' in \emph{2023 IEEE International Conference on Robotics and Automation (ICRA)}.\hskip 1em plus 0.5em minus 0.4em\relax IEEE, 2023, pp. 3592--3599.

\bibitem{desolneux2000meaningful}
A.~Desolneux, L.~Moisan, and J.-M. Morel, ``Meaningful alignments,'' \emph{International journal of computer vision}, vol.~40, pp. 7--23, 2000.

\bibitem{bammey2023contrario}
Q.~Bammey, ``A contrario mosaic analysis for image forensics,'' in \emph{International Conference on Advanced Concepts for Intelligent Vision Systems}.\hskip 1em plus 0.5em minus 0.4em\relax Springer, 2023, pp. 222--234.

\bibitem{von2008lsd}
R.~G. Von~Gioi, J.~Jakubowicz, J.-M. Morel, and G.~Randall, ``Lsd: A fast line segment detector with a false detection control,'' \emph{IEEE transactions on pattern analysis and machine intelligence}, vol.~32, no.~4, pp. 722--732, 2008.

\bibitem{xia2014accurate}
G.-S. Xia, J.~Delon, and Y.~Gousseau, ``Accurate junction detection and characterization in natural images,'' \emph{International journal of computer vision}, vol. 106, pp. 31--56, 2014.

\bibitem{feuge2023contrario}
B.~G. Feuge-Miller, M.~K. Jah, A.~T. Karra, S.~Iyer, and D.~Kucharski, ``A-contrario detection and tracking from optical telescope data,'' \emph{Acta Astronautica}, vol. 210, pp. 129--140, 2023.

\bibitem{DBLP:journals/corr/abs-1803-01711}
\BIBentryALTinterwordspacing
A.~Flenner, L.~Peterson, J.~Bunk, T.~M. Mohammed, L.~Nataraj, and B.~S. Manjunath, ``Resampling forgery detection using deep learning and a-contrario analysis,'' \emph{CoRR}, vol. abs/1803.01711, 2018. [Online]. Available: \url{http://arxiv.org/abs/1803.01711}
\BIBentrySTDinterwordspacing

\bibitem{li2018megadepth}
Z.~Li and N.~Snavely, ``Megadepth: Learning single-view depth prediction from internet photos,'' in \emph{Proceedings of the IEEE conference on computer vision and pattern recognition}, 2018, pp. 2041--2050.

\bibitem{dai2017scannet}
A.~Dai, A.~X. Chang, M.~Savva, M.~Halber, T.~Funkhouser, and M.~Nie{\ss}ner, ``Scannet: Richly-annotated 3d reconstructions of indoor scenes,'' in \emph{Proceedings of the IEEE conference on computer vision and pattern recognition}, 2017, pp. 5828--5839.

\bibitem{anguelov2010google}
D.~Anguelov, C.~Dulong, D.~Filip, C.~Frueh, S.~Lafon, R.~Lyon, A.~Ogale, L.~Vincent, and J.~Weaver, ``Google street view: Capturing the world at street level,'' \emph{Computer}, vol.~43, no.~6, pp. 32--38, 2010.

\bibitem{desolneux2016contrario}
A.~Desolneux, ``When the a contrario approach becomes generative,'' \emph{International Journal of Computer Vision}, vol. 116, no.~1, pp. 46--65, 2016.

\bibitem{wan2019contrario}
Y.~Wan, Y.~Zhang, and X.~Liu, ``An a-contrario method of mismatch detection for two-view pushbroom satellite images,'' \emph{ISPRS Journal of Photogrammetry and Remote Sensing}, vol. 153, pp. 123--136, 2019.

\bibitem{robin2010contrario}
A.~Robin, L.~Moisan, and S.~Le~H{\'e}garat-Mascle, ``An a-contrario approach for subpixel change detection in satellite imagery,'' \emph{IEEE Transactions on pattern analysis and machine intelligence}, vol.~32, no.~11, pp. 1977--1993, 2010.

\bibitem{deep_nfa}
A.~Ciocarlan, S.~{Le Hégarat-Mascle}, S.~Lefebvre, and A.~Woiselle, ``Deep-nfa: A deep a contrario framework for tiny object detection,'' \emph{Pattern Recognition}, vol. 150, p. 110312, 2024.

\end{thebibliography}

\end{document}